\documentclass{article} 
\usepackage{iclr2026_conference,times}

\usepackage{amsmath,amsfonts,bm}

\def\eqref#1{equation~\ref{#1}}

\def\1{\bm{1}}

\def\vc{{\bm{c}}}

\def\vo{{\bm{o}}}

\def\vx{{\bm{x}}}

\def\vz{{\bm{z}}}

\def\mH{{\bm{H}}}
\def\mI{{\bm{I}}}

\def\mP{{\bm{P}}}

\DeclareMathAlphabet{\mathsfit}{\encodingdefault}{\sfdefault}{m}{sl}
\SetMathAlphabet{\mathsfit}{bold}{\encodingdefault}{\sfdefault}{bx}{n}

\newcommand{\R}{\mathbb{R}}

\newcommand{\KL}{D_{\mathrm{KL}}}

\usepackage{hyperref}
\usepackage{url}
\usepackage{xspace}
\usepackage[most]{tcolorbox}
\usepackage{wrapfig}
\usepackage{booktabs}
\usepackage{graphicx}
\usepackage{subcaption}
\usepackage{multirow}
\usepackage[capitalize]{cleveref}
\usepackage{hyperref}

\makeatletter
\DeclareRobustCommand\onedot{\futurelet\@let@token\@onedot}
\def\@onedot{\ifx\@let@token.\else.\null\fi\xspace}

\def\eg{\emph{e.g}\onedot} 
\def\ie{\emph{i.e}\onedot}

\makeatother

\title{ImageDoctor: Diagnosing Text-to-Image Generation via Grounded Image Reasoning}

\author{\textbf{Yuxiang Guo}{\footnotesize $^{1,2}$}\thanks{Equal Contribution} \;
 \textbf{Jiang Liu}{\footnotesize $^{2*}$}
 \;
 \textbf{Ze Wang}{\footnotesize $^{2}$}
 \;
 \textbf{Hao Chen}{\footnotesize $^2$}
  \;
 \textbf{Ximeng Sun}{\footnotesize $^2$}
 \\
 \textbf{Yang Zhao}{\footnotesize $^{1}$}
 \;
  \textbf{Jialian Wu}{\footnotesize $^{2}$}
  \;
  \textbf{Xiaodong Yu}{\footnotesize $^2$}
  \;
  \textbf{Zicheng Liu}{\footnotesize $^{2}$}
  \;
  \textbf{Emad Barsoum}{\footnotesize $^{2}$}
  \\[2pt]
{\footnotesize $^1$}Johns Hopkins University \;
{\footnotesize $^2$}AMD\; \\
{\url{https://image-doctor.github.io/}}
}

\iclrfinalcopy 
\begin{document}

\maketitle

\begin{abstract}

\looseness -1 The rapid advancement of text-to-image (T2I) models has increased the need for reliable human preference modeling, a demand further amplified by recent progress in reinforcement learning for preference alignment. However, existing approaches typically quantify the quality of a generated image using a single scalar, limiting their ability to provide comprehensive and interpretable feedback on image quality. To address this, we introduce ImageDoctor, a unified multi-aspect T2I model evaluation framework that assesses image quality across four complementary dimensions: plausibility, semantic alignment, aesthetics, and overall quality. ImageDoctor also provides pixel-level flaw indicators in the form of heatmaps, which highlight misaligned or implausible regions, and can be used as a dense reward for T2I model preference alignment. Inspired by the diagnostic process, we improve the detail sensitivity and reasoning capability of ImageDoctor by introducing a \textit{``look-think-predict"} paradigm, where the model first localizes potential flaws, then generates reasoning, and finally concludes the evaluation with quantitative scores. Built on top of a vision-language model and trained through a combination of supervised fine-tuning and reinforcement learning, ImageDoctor demonstrates strong alignment with human preference across multiple datasets, establishing its effectiveness as an evaluation metric. Furthermore, when used as a reward model for preference tuning, ImageDoctor significantly improves generation quality—achieving an improvement of 10\% over scalar-based reward models.

\end{abstract}

\section{introduction}

With the rapid evolution of text-to-image (T2I) architectures~\citep{goodfellow2014generative, croitoru2023diffusion, gu2023matryoshka, xie2024show, tian2024visual, wang2025instella}, the quality of generated images has advanced significantly. Modern T2I systems can now produce highly realistic outputs that closely follow textual instructions, enabling a wide range of applications in areas such as art, design, and entertainment. These advances, however, give rise to a critical question: \emph{how to reliably evaluate the quality of generated images that may suffer from poor instruction adherence, low aesthetic quality, or counterintuitive artifacts}. Furthermore, with the rise of reinforcement learning~\citep{liu2025flow, xue2025dancegrpo} and test-time scaling~\citep{guo2025can,ma2025inference}, evaluators play an increasingly important role: not only can they serve as reward functions or verifiers to measure quality, but they are also expected to provide actionable feedback to improve generation.

Current human preference models, such as HPS~\citep{wu2023human1}, ImageReward~\citep{xu2023imagereward}, and PickScore~\citep{kirstain2023pick}, mostly predict a single scalar score of quality. However, compressing the evaluation into a single scalar is often insufficient to capture the detailed flaws of the generated images. For instance, two images may receive the same score, yet differ substantially: one may be aesthetically pleasing but poorly aligned with the prompt, while the other may follow the prompt faithfully but contain unrealistic artifacts. Relying solely on a single score cannot disentangle these factors, limiting both the interpretability and the usefulness of the feedback for guiding model improvement. Moreover, existing evaluators can only provide an overall judgment of image quality but lack spatially grounded feedback, \ie, they cannot identify \textit{where} in the image the problems occur.  In practice, many T2I failures stem from partial prompt adherence: while the majority of the prompt may be satisfied, fine-grained details are often missing or incorrect. This absence of localization further reduces the interpretability and actionability of these evaluators, especially when used as reward functions.
These challenges highlight the need for evaluators that can provide comprehensive feedback, offering both multi-dimensional quality scores and localized diagnostics—much like a doctor diagnosing the problems in an image.

\begin{figure}
    \centering
    \includegraphics[width=\linewidth]{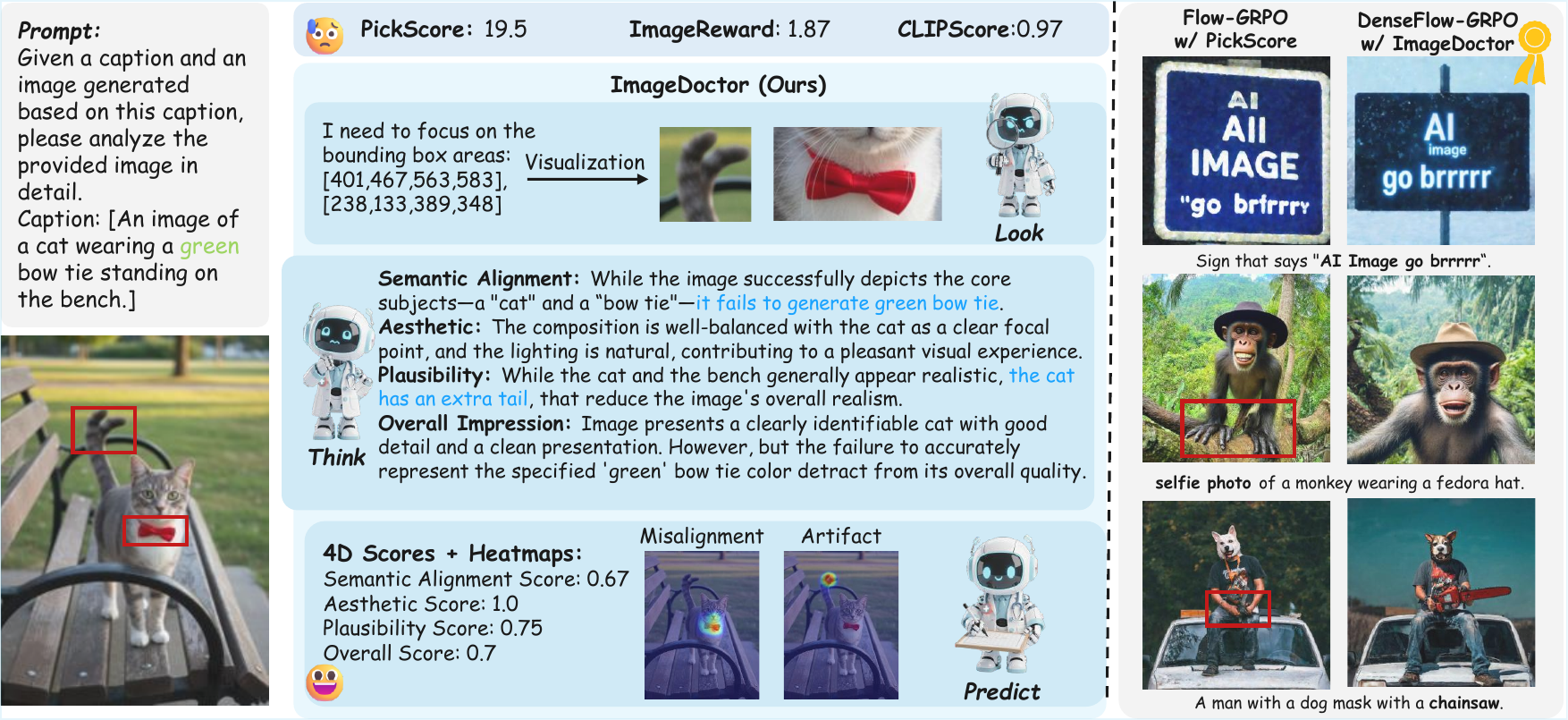}
    \caption{\textbf{Comparison between ImageDoctor and scalar-based reward functions.} \textit{Left}: ImageDoctor follows a \textit{``look-think-predict"} paradigm, providing rich feedback with four-dimensional scores and heatmaps that highlight misalignment and artifact locations. \textit{Right}: Leveraging this fine-grained feedback, DenseFlow-GRPO (Sec. \ref{sec:denseflow-grpo}) generates images with more faithful and realistic local details, outperforming Flow-GRPO, which relies on the scalar-based reward PickScore. }
    \label{fig:teaser}
    \vspace{-15pt}
\end{figure}

In this work, we propose \textbf{ImageDoctor}, a unified evaluation framework that produces holistic scoring and spatially grounded feedback in the form of artifact and misalignment heatmaps. Steering the reasoning strengths and commonsense knowledge of multi-modal large language models (MLLMs), ImageDoctor is built on a fine-tuned MLLM backbone to achieve a deep joint understanding of images and prompts. To achieve flaw localization, we introduce a lightweight heatmap decoder that produces the heatmaps highlighting misalignment and artifact locations conditioned on the input prompt, image features, and the response generated by the MLLM.
Inspired by the process of medical diagnosis, we further propose a \textit{``look-think-predict"} paradigm as shown in~\cref{fig:teaser}. Before final judgment, ImageDoctor performs grounded image reasoning, which consists of two steps. First, it pinpoints potential flawed regions that require closer attention in the image (\textit{``look"}). Then, it analyzes these regions by integrating the localized visual evidence with contextual understanding, generating structured reasoning that evaluates the image from multiple aspects (\textit{``think"}). We incentivize this grounded image reasoning capability with cold start and reinforcement finetuning.

Reinforcement learning from human feedback (RLHF)~\citep{xue2025dancegrpo, liu2025flow} has been proven effective in enhancing both image quality and text-image alignment for T2I generation. Nevertheless, current RLHF approaches, such as Flow-GRPO~\citep{liu2025flow}, rely solely on sparse \textit{image-level} rewards, which overlook spatially localized feedback and thus fail to provide fine-grained guidance during training. To address this limitation, we introduce \textbf{DenseFlow-GRPO}, a new RLHF framework that enhances T2I models with both \textit{image-level} and \textit{pixel-level} dense reward signals. By leveraging the rich diagnostic feedback from ImageDoctor, DenseFlow-GRPO delivers more precise and spatially aligned supervision, enabling T2I models to learn not only what constitutes a good image globally, but also how to refine local regions in a fine-grained manner.

Our experiments demonstrate that ImageDoctor achieves state-of-the-art alignment with human judgments, substantially improving score prediction accuracy across all dimensions (average PLCC 0.741 vs. 0.586 of the previous best on RichHF-18K). Beyond serving as a metric, ImageDoctor generalizes well to downstream applications: as a verifier, it reliably selects higher-quality generations in test-time scaling; as a reward model, it drives consistent gains in reinforcement learning. In particular, integrating ImageDoctor into Flow-GRPO yields superior preference alignment, and further utilizing the dense heatmap feedback in DenseFlow-GRPO achieves the strongest improvements and delivers images with more faithful local details.

The main contributions of this paper are summarized below:

\begin{itemize}

    \item We propose ImageDoctor, a unified T2I evaluation model that produces dense feedback, including multi-aspect scores and heatmaps localizing flaws, enabling interpretable and fine-grained assessment.
    \item We introduce a \textit{``look-think-predict"} paradigm that equips ImageDoctor with structured reasoning by integrating visual grounding and textual analysis. ImageDoctor is further refined through reinforcement finetuning with tailored reward functions, enhancing adherence to human preferences while ensuring spatially grounded reasoning.
    \item We present DenseFlow-GRPO, a novel T2I reinforcement learning method that incorporates ImageDoctor’s dense spatial feedback into the reward signal, providing region-aware supervision and leading to more robust improvements in image generation.
    \item Extensive experiments on human preference datasets demonstrate the ImageDoctor's superior alignment with human preference, and we further validate its effectiveness by applying it to downstream tasks, serving as a verifier and a reward model.
\end{itemize}

\section{Related work}
\vspace{-6pt}
\subsection{Text-to-image Generation}

Text-to-image (T2I) generation is a core task in generative modeling. It aims to synthesize semantically aligned images from natural language prompts, while balance the aesthetic quality and plausibility.
Early approaches based on Generative Adversarial Networks (GANs)~\citep{goodfellow2014generative} and Variational Auto-Encoders (VAEs)~\citep{kingma2013auto} demonstrated feasibility but were limited by low diversity and coarse details.
More recently, diffusion models~\citep{ho2020denoising, nichol2021glide, rombach2022high} have emerged as a dominant paradigm, achieving significant gains in image quality and diversity. Flow-based models~\citep{zhao2024flowturbo,esser2024scaling,flux2024} provide another class of likelihood-based generative models, relying on stochastic denoising steps, thereby enabling efficient sampling and reducing inference overhead. In addition, autoregressive models~\citep{tian2024visual, xie2024show} are gaining attention for their compositionality and controllability, bridging vision and language more effectively.
\vspace{-6pt}

\subsection{T2I Evaluation Models}
With the rapid progress in T2I generation, evaluation models have also advanced, though the task remains highly challenging. 
CLIPScore~\citep{hessel2021clipscore} was one of the earliest automatic metrics that leverages pretrained CLIP to compute the similarity between the generated image and its prompt. PickScore~\citep{kirstain2023pick} and ImageReward~\citep{xu2023imagereward} fine-tune CLIP~\citep{radford2021learning} and BLIP~\citep{li2022blip}, respectively, on large-scale human preference datasets, significantly improving alignment with subjective human judgments. 
The Human Preference Score (HPS) series~\citep{wu2023human1, wu2023human2, ma2025hpsv3} expand the scale of annotations to enhance preference alignment.
Recently, UnifiedReward-think~\citep{unifiedreward-think} and VisualQuality-R1~\citep{tian2024visual} explored reinforcement learning for evaluation model training for image quality score prediction. 
RichHF~\citep{liang2024rich} and HELM~\citep{lee2023holistic} attempt to broaden evaluation by considering multiple dimensions, moving beyond single-score preference modeling.

\section{ImageDoctor}

\begin{figure}
    \centering
    \includegraphics[width=.9\linewidth]{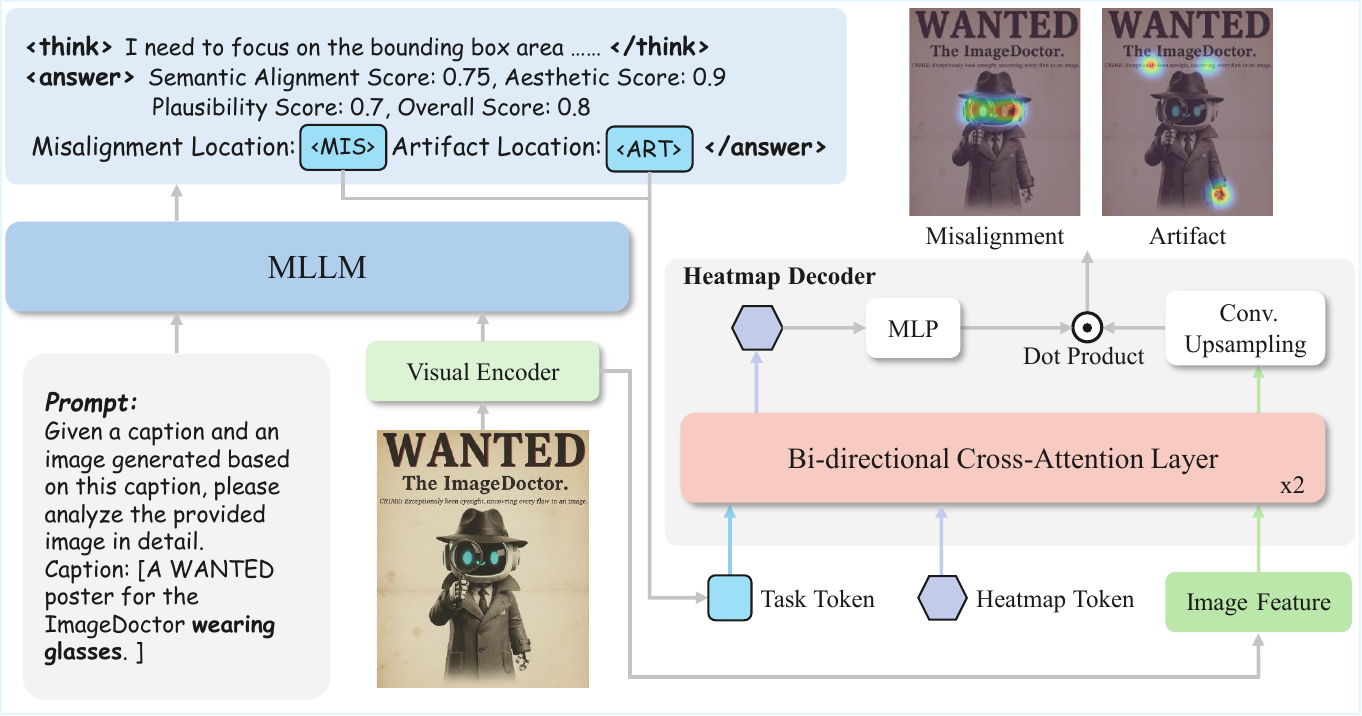}
    \caption{\textbf{ImageDoctor architecture.} Given a prompt-image pair, the MLLM follows a \textit{``look-think-predict"} paradigm for T2I evaluation by localizing potential flaw regions, analyzing them, and generating holistic scores and special task tokens. The task token, with a learned heatmap token and image features are fed into the heatmap decoder to produce the misalignment and artifact heatmaps. }
    \label{fig:Pipeline}
    \vspace{-10pt}
\end{figure}

\vspace{-6pt}
\looseness -1 ImageDoctor aims to provide rich, interpretable, and accurate diagnoses for T2I generation. To this end, we design a novel unified model architecture to generate both \textit{image-level} scores and \textit{pixel-level} heatmap evaluations leveraging the strong image and text understanding capabilities of multimodal large language models (MLLMs) (Sec. \ref{sec:model}). To generate interpretable and accurate evaluations, we propose a {\textit{``look-think-predict"}} paradigm, where the model first identifies possible local flaw regions and generates explicit reasoning about image details before providing final evaluations (Sec. \ref{sec:think}). 

\vspace{-6pt}

\subsection{Model Design}
\vspace{-6pt}

\label{sec:model}

\noindent \paragraph{\textbf{Overview.}} The overall pipeline of ImageDoctor is shown in Fig.~\ref{fig:Pipeline}. The input text prompt $\mP$ and the corresponding image $\mI \in \R^{H \times W}$ are passed into the MLLM backbone. ImageDoctor reasons about image details and image-text semantics to produce both holistic scores and localized diagnostic signals. Specifically, it outputs four scalar scores, \ie, semantic alignment, aesthetics, plausibility, and overall scores $s_{d}, \forall d \in \{\text{align, aesth, plau, over}\}$, which evaluate image quality from different aspects. In addition, it provides localized feedback by marking image regions that are implausible or misaligned with the text through the artifact and misalignment heatmaps $\mH_d \in \mathbb{R}^{H \times W}, \forall d \in \{\text{art, mis}\}$ generated by the heatmap decoder.

\vspace{-6pt}

\noindent \paragraph{\textbf{Heatmap Decoder.}}  While scalar scores can be directly predicted via the text output of the MLLM backbone, generating pixel-wise heatmaps requires a unified model that supports both text and image outputs. To enable this, we design a lightweight heatmap decoder. The decoder takes the image features extracted by the visual encoder together with a learned heatmap token and a task token $\mathbf{t}\in\{\texttt{<ART>}, \texttt{<MIS>} \}$ representing the artifact and the misalignment tokens, respectively. The task tokens are generated by the MLLM backbone and fused with the input image, text prompt, and reasoning chains to guide accurate heatmap prediction. Inspired by the SAM mask decoder~\citep{kirillov2023segment}, we adopt a bi-directional cross-attention design to fuse the token and image embeddings. The updated image features are passed through a series of convolution upsampling layers to upscale to the original image size, from which the updated heatmap token are used to dynamically predict the heatmap. The detailed architecture of the heatmap decoder can be found in~\cref{sec:decoder}.

\vspace{-6pt}
\subsection{Grounded Image Reasoning}
\label{sec:think}

As shown in~\cref{fig:teaser}, ImageDoctor adopts a \textbf{\textit{``look-think-predict"}} paradigm to generate its evaluations for a given image and text prompt. Instead of directly generating the final prediction, it first localizes potential flaw regions by predicting the flaw region bounding boxes (\textit{\textbf{``look"}}), then analyzes and reasons about these flaws and overall image quality (\textit{\textbf{``think"}}), and finally produces a conclusive judgment (\textit{\textbf{``predict"}}), mimicking the image evaluation process of human. To enable the grounded image reasoning capability, we design a two-phase training pipeline. In the {cold start} phase, we conduct supervised fine-tuning to teach the model to predict image scores and heatmaps in the \textit{``look-think-predict"} reasoning format. In the second phase, ImageDoctor adopts online {reinforcement fine-tuning} with Group Relative Policy Optimization (GRPO)~\citep{shao2024deepseekmath} to further incentivize the reasoning ability to generate rich and reliable image feedback.

\vspace{-6pt}

\subsubsection{Cold Start}
Since MLLMs are designed for general image understanding, in the first cold start stage, we first finetune the MLLM backbone on the image evaluation task by training the model to directly predict the image evaluation scores. In the second cold start stage, we train ImageDoctor on chain-of-thought (CoT) data to learn the \textit{``look-think-predict"} reasoning process for both score and heatmap predictions. To prepare the CoT data, we first detect the highlighted regions in the ground-truth artifact and misalignment heatmaps to generate flaw region bounding boxes. Then, we employ Gemini 2.5 Flash~\citep{comanici2025gemini} with carefully designed prompts to produce detailed reasoning between the image and corresponding human annotations. Finally, we organize the bounding boxes, reasoning traces, and ground-truth human annotations into the \textit{``look-think-predict"} CoT format. The details can be found in~\cref{sec:reasoning_generation}. The ImageDoctor model $\theta$ is optimized by:

\begin{equation}
    \mathcal{L} = - \sum_{i} \log p_\theta(\vz_i \mid \vz_{<i}, \mI, \mP)+ \sum_{d} \| \mH_d - \tilde{\mH}_d \|_2^2,
\end{equation}
where $\vz$ is the CoT reasoning text, $\mI$ and $\mP$ are the input image and text prompts, $\mH_d$ and $\tilde{\mH}_d$ are the ground-truth and predicted heatmaps, respectively.

\vspace{-6pt}
\subsubsection{Reinforcement Finetuning}
\label{sec:reinforcement}
After cold start, we further perform reinforcement finetuning (RFT) with GRPO~\citep{shao2024deepseekmath} to enhance the reasoning ability of ImageDoctor. Given a pair of input $(\mI, \mP)$, ImageDoctor as the policy model $\pi_{\theta}$, generates a group of $N$ candidate responses $\{\vo^{1},\ldots,\vo^{N}\}$. For each response $\vo^{i}$, we compute a reward score $\mathcal{R}^{i}$ using a combination of reward functions. The rewards are normalized within the group to compute the group-normalized  advantage. RFT allows the model to explore diverse reasoning paths, directing it toward reasoning trajectories with high reward signals and enhancing its generalization capability. The detailed GRPO formulation can be found in~\cref{sec:GRPO}.

We design a suite of verifiable rewards to encourage the model to focus on the correct flaw regions, produce accurate evaluation scores, and generate precise heatmaps, including a grounding reward ($\mathcal{R}_G$), a score reward ($\mathcal{R}_S$) and a heatmap reward ($\mathcal{R}_H$).

\looseness -1 \noindent {\textbf{Grounding Reward}} ($\mathcal{R}_G$) aims to evaluate whether the model can accurately locate the flaw regions in an image. The model should ideally generate a compact set of bounding boxes, both in number and area, that effectively cover the potential flaw regions. The grounding reward $\mathcal{R}_G$ has three complementary components: \textbf{1) Completeness.} The union of all bounding boxes should adequately cover the entire highlighted area in the artifact and misalignment heatmaps. We compute the ratio between the area covered by the union of all bounding boxes and the total intensity of the heatmaps. \noindent\textbf{2) Compactness.} Each bounding box should only cover flaw regions with minimal normal regions. We compute the average heatmap intensity within each predicted bounding box and then take the mean across all boxes, yielding higher rewards for bounding boxes with less normal regions. \textbf{3) Uniqueness.} The model should not predict redundant bounding boxes, and thus the overlap between any pair of boxes should be minimized. We measure the Intersection over Union (IoU) between each pair of bounding boxes and apply a penalty for large overlaps. Implementation details are in~\cref{sec:grounding_reward}.

\noindent {\textbf{Score Reward}} ($\mathcal{R}_s$) evaluates how well the predicted scores $\tilde{s}_d$ align with ground-truth human scores ${s}_d$. We use $\ell_1$ distance and design $\mathcal{R}_S=\sum_{d}1- \|s_d-\tilde{s}_d\|_1$, which encourages the model to produce score predictions that are close to human judgments.

\noindent {\textbf{Heatmap Reward}} ($\mathcal{R}_H$) measures the similarity between predicted heatmaps $\tilde{\mH}_d$ and human annotated heatmaps $\mH_d$. We use $\ell_2$ distance and define $\mathcal{R}_H=\sum_{d} 1- \| \mH_d - \tilde{\mH}_d \|_2^2.$ This formulation assigns higher rewards when the predicted maps closely match the annotations, thereby encouraging the model to produce precise and sharp flaw localization heatmaps.

The total reward is the combination of the three rewards: $\mathcal{R}=\mathcal{R}_G+\mathcal{R}_S+\mathcal{R}_H$.

\vspace{-6pt}

\section{DenseFlow-GRPO: ImageDoctor as Dense Reward}
\label{sec:denseflow-grpo}

Reinforcement learning from human feedback (RLHF)~\citep{xue2025dancegrpo, liu2025flow} has demonstrated great success in improving image quality and image-text alignment for T2I generation. However, existing RLHF methods such as Flow-GRPO~\citep{liu2025flow} adopt an \textit{image-level} formulation without fine-grained supervision. Specifically, given a prompt \(\vc\), the flow model \(p_{\phi}\) samples a group of \(G\) individual images \(\{\vx^i_0\}_{i=1}^G\) and the corresponding reverse-time trajectories \(\{(\vx^i_T, \vx^i_{T -1}, \cdots, \vx^i_0)\}_{i=1}^G\). 
Flow-GRPO optimizes the flow model by maximizing the following: 
\begin{equation}
\mathcal{J}_\text{Flow-GRPO}(\phi) =
\frac{1}{G}\sum_{i=1}^{G} \frac{1}{T}\sum_{t=0}^{T-1} \Bigg( 
\min \Big( r^i_t(\phi) \hat{A}^i_t,  
\ \text{clip} \Big( r^i_t(\phi), 1 - \varepsilon, 1 + \varepsilon \Big) \hat{A}^i_t \Big)
- \beta \KL(p_{\phi} || p_{\phi_{\text{ref}}}) 
\Bigg), 
\label{eq:densegrpoloss}
\end{equation}
where the likelihood ratio $ r^i_t(\phi)$ and normalized advantage $\hat{A}^i_t$ of the \(i\)-th image are computed as:

\begin{equation}
 r^i_t(\phi) =
\frac{p_{\phi}(\vx^i_{t-1} \mid \vx^i_t, \vc)}{p_{\phi_{\text{old}}}(\vx^i_{t-1} \mid \vx^i_t, \vc)}, \ \    \hat{A}^i_t = \frac{R(\vx^i_0, \vc) - \text{mean}(\{R(\vx^i_0, \vc)\}_{i=1}^G)}{\text{std}(\{R(\vx^i_0, \vc)\}_{i=1}^G)}.
\label{eq:grpo_r_a}
\end{equation}

In Eq.~(\ref{eq:grpo_r_a}), both $r^i_t(\phi)$ and reward $R$ are computed on the image level. The reward signal is applied uniformly across all pixels in the image, treating every region equally, regardless of its quality. Finer-grained supervision is more desirable, as it allows low-quality regions to be penalized more, while encouraging high-quality areas. 
To fill this gap, we propose \textbf{DenseFlow-GRPO}, which enables both \textit{image-level} and \textit{pixel-level} fine-grained dense reward signals for flow model RL training, leveraging the rich image feedback generated by ImageDoctor. We first reformulate the likelihood ratio at each trajectory step to allow pixel-wise advantage customization: 
\begin{equation}
s_{t}^i(\phi,h,w) = \mathrm{sg}\!\left[ r^i_t(\phi)\right] \cdot 
\frac{p_{\phi}(\vx^i_{t-1} \mid \vx^i_t, \vc)_{h,w}}
{\mathrm{sg}\!\left[p_{\phi}(\vx^i_{t-1} \mid \vx^i_t, \vc)_{h,w}\right]},
\label{eq:pixel_likelihood}
\end{equation}
where $p_{\phi}(\vx^i_{t-1} \mid \vx^i_t, \vc)_{h,w}$ is the pixel-wise likelihood, $h,w$ denote the pixel location, and $\mathrm{sg}\!\left[\cdot\right]$ is the stop-gradient operation that only takes the numerical value, corresponding to \texttt{detach} in PyTorch. We can then apply the dense pixel-wise advantage that combines \textit{image-level} reward $R$ and \textit{pixel-level} reward $R_P$:
\begin{equation}
    \hat{A}^i_t(h,w) = \frac{R_D(\vx^i_0, \vc, h, w) - \text{mean}(\{R_D(\vx^i_0, \vc, h, w)\}_{i=1}^G)}{\text{std}(\{R_D(\vx^i_0, \vc, h, w)\}_{i=1}^G)}.
\end{equation}
where $R_D(\vx^i_0, \vc, h, w) = R(\vx^i_0, \vc) + R_P(\vx^i_0, \vc, h, w)$ is the dense reward function.

The DenseFlow-GRPO objective is defined as:

\begin{equation}
\mathcal{J_\text{Dense}}(\phi) =
\frac{1}{GTHW}\sum_{i,t,h,w}\Bigg( 
\min \Big( s^i_t(\phi,h,w) \hat{A}^i_t(h,w),  
\ \text{clip} \Big( s^i_t(\phi,h,w), 1 - \varepsilon, 1 + \varepsilon \Big) \hat{A}^i_t(h,w) \Big)
\Bigg), 
\label{eq:grpoloss}
\end{equation}

where we omit the KL regularization term for brevity. Note that in Eq.~(\ref{eq:pixel_likelihood}), $s_{t}^i(\phi,h,w)$ is numerically equal to $r^i_t(\phi)$ but allows the pixel-wise advantage to backpropagate to the local image regions through $p_\phi(\cdot)_{h,w}$. This is similar to the GSPO-token~\citep{zheng2025group} formulation, and we find it more stable than directly computing the pixel-wise likelihood ratio.

\vspace{-6pt}
\section{Experiments}

\vspace{-6pt}
\subsection{Experimental Setup}
\noindent\textbf{Datasets.} 
We train and evaluate ImageDoctor on the \textit{RichHF-18K}~\citep{liang2024rich} dataset. RichHF-18K is a subset of Pick-a-Pic~\citep{kirstain2023pick}, consisting of 16K training samples, 1K validation samples, and 1K test samples. For each text-image pair, it provides two heatmaps and four fine-grained scores annotated by a total of 27 annotators. To further assess the generalizability of ImageDoctor, we also test it on the \textit{GenAI-Bench}~\citep{li2024evaluating} and \textit{TIFA}~\citep{hu2023tifa}.

\noindent\textbf{Evaluation Metrics.} We follow the official evaluation protocols of the datasets. For score prediction, we employ Pearson Linear Correlation Coefficient (PLCC) and Spearman Rank Correlation Coefficient (SRCC), which measure how well the predicted scores correlate with human annotations. For heatmap prediction, we report the Mean Squared Error (MSE) between predictions and ground truth. Additionally, we adopt standard heatmap metrics~\citep{liang2024rich} including KL Divergence (KLD), Similarity (SIM), and Correlation Coefficient (CC), providing a comprehensive assessment of spatial prediction quality.

\noindent\textbf{Implementation Details} All experiments are conducted on four AMD MI250 GPUs. We adopt Qwen2.5-VL-3B~\citep{Qwen2.5-VL} as the MLLM backbone. Training is performed for 5 epochs in cold start stage 1 and 3 epochs in stage 2. We train for 400 steps for RFT. Learning rates are set to $1\times10^{-5}$ for cold start and $1\times10^{-6}$ for RFT. Training images are resized to $512 \times 512$.

\vspace{-5pt}

\subsection{Main Results}
\vspace{-3pt}

\begin{table}[t]
    \setlength{\tabcolsep}{4pt}
    \centering
    \caption{Performance comparison of score prediction on RichHF-18K.}
     \resizebox{\textwidth}{!}{
    \begin{tabular}{l|cccccccc|cc}
        \toprule
        \multirow{2}{*}{Method} & \multicolumn{2}{c}{Plausibility} & \multicolumn{2}{c}{Aesthetics} & \multicolumn{2}{c}{Semantic Alignment} & \multicolumn{2}{c}{Overall} & \multicolumn{2}{|c}{Average} \\
        \cmidrule(lr){2-3} \cmidrule(lr){4-5} \cmidrule(lr){6-7} \cmidrule(lr){8-9} \cmidrule(lr){10-11}
        & PLCC $\uparrow$ & SRCC $\uparrow$ & PLCC $\uparrow$ & SRCC $\uparrow$ & PLCC $\uparrow$ & SRCC $\uparrow$ & PLCC $\uparrow$ & SRCC $\uparrow$ & PLCC $\uparrow$ & SRCC $\uparrow$ \\
        \midrule
        ResNet-50~\citep{he2016deep} & 0.495 & 0.487 & 0.370 & 0.363 & 0.108 & 0.119 & 0.337 & 0.308 & 0.328 & 0.319 \\
        CLIP~\citep{radford2021learning}  & 0.390 & 0.378 & 0.357 & 0.360 & 0.398 & 0.390 & 0.353 & 0.352 &0.374 & 0.370 \\
        PickScore~\citep{kirstain2023pick} & 0.010 & 0.028 & 0.131 & 0.140 & 0.346 & 0.340 & 0.202 & 0.226 & 0.172 & 0.183 \\
        RichHF~\citep{liang2024rich} & 0.693 & 0.681 & 0.600 & 0.589 & 0.474 & 0.496 & 0.580 & 0.562 & 0.586 & 0.582\\
        \midrule
        \textbf{ImageDoctor (Ours)} & \textbf{0.727} &\textbf{ 0.711} & \textbf{0.681} & \textbf{0.662} & \textbf{0.808} &  \textbf{0.799} & \textbf{0.745} & \textbf{0.725}  &\textbf{0.741} &\textbf{ 0.724} \\
        \bottomrule
    \end{tabular}
    }

    \label{tab:score_prediction}
\end{table}

\begin{table}[t]
  \centering
  \caption{Performance comparison of heatmap prediction on RichHF-18K.}
  \setlength{\tabcolsep}{5pt}
  \renewcommand{\arraystretch}{0.98}
  \resizebox{\textwidth}{!}{
  \begin{tabular}{l|ccccc|ccccc} 
    \toprule
    \multirow{4}{*}{Method} & \multicolumn{5}{c|}{Artifact} & \multicolumn{5}{c}{Misalignment} \\
    \cmidrule(lr){2-11}
    & \multicolumn{1}{c|}{All data} & \multicolumn{1}{c|}{$GT=0$} &  \multicolumn{3}{c|}{$GT>0$}  & \multicolumn{1}{c|}{All data} & \multicolumn{1}{c|}{$GT=0$} &  \multicolumn{3}{c}{$GT>0$}\\
    \cmidrule(lr){2-11}
    &\multicolumn{1}{c|}{ MSE $\downarrow$} &\multicolumn{1}{c|}{MSE $\downarrow$} & CC $\uparrow$ & KLD $\downarrow$ & SIM $\uparrow$ &\multicolumn{1}{|c|}{ MSE $\downarrow$} &\multicolumn{1}{c|}{MSE $\downarrow$} & CC $\uparrow$ & KLD $\downarrow$ & SIM $\uparrow$ \\
    \midrule
    ResNet-50~\citep{he2016deep}  & 0.00996 & 0.00093 & 0.506 & 1.669 & 0.338 & - & - & - & - & -  \\
    CLIP Gradient~\citep{simonyan2013deep} & - & - & - & - & - & 0.00817 & 0.00551 & 0.015 & 3.844 & 0.041 \\
    
    RichHF~\citep{liang2024rich} & 0.00920 & 0.00095 & 0.556 & 1.652 & 0.409 & 0.00304 & 0.00006 & 0.212 & 2.933 & 0.106\\
    \midrule
   \textbf{ImageDoctor (Ours)} & \textbf{0.00891} & \textbf{0.00076} & \textbf{0.571} & \textbf{1.477} & \textbf{0.412} & \textbf{0.00299} & \textbf{0.00003} & \textbf{0.225} & \textbf{2.863} & \textbf{0.108}\\
    \bottomrule 
  \end{tabular}}
  \vspace{-10pt}
  \label{tab:performance_heatmap}
\end{table}

\noindent \textbf{Results on RichHF-18K.} \Cref{tab:score_prediction} shows the score prediction results across four dimensions on RichHF-18K. We compare with ResNet-50 ~\citep{he2016deep}, CLIP~\citep{radford2021learning} models fine-tuned on the RichHF-18K dataset, as well as the off-the-shelf PickScore~\citep{kirstain2023pick} model. In addition, we compare with the RichHF model~\citep{liang2024rich}, which is trained on RichHF-18k and is able to generate both score and heatmap predictions. ImageDoctor achieves the best performance across all dimensions, substantially improving semantic alignment (PLCC: 0.808 vs. 0.474) and raising the average PLCC from 0.586 to 0.741 compared to the previous best method RichHF, demonstrating much stronger correlation with human judgment. These gains also extend to heatmap prediction results in \Cref{tab:performance_heatmap}, where ImageDoctor achieves the best performance, highlighting its ability to precisely localize flaws in generated images.

\noindent \textbf{Results on GenAI-Bench and TIFA.} \Cref{tab:cross_domain} presents the results on the GenAI-Bench and TIFA datasets. To assess the generalizability of ImageDoctor, we evaluate the model trained solely on RichHF-18K without any fine-tuning on these two benchmarks. Despite differences in image sources and the inherent subjectivity of annotators, ImageDoctor consistently outperforms previous human preference models, achieving higher correlations with human annotations.

\begin{table}[t]
    \centering
    \begin{minipage}[c]{0.4\textwidth}
    \setlength{\tabcolsep}{4pt}
        \centering
        \vspace{0pt}
            \caption{Quantitative comparison on the GenAI-Bench and TIFA datasets.}
    \resizebox{\textwidth}{!}{
    \begin{tabular}{lcccccccccccc}
        \toprule
        \multirow{2}{*}{Method}  & \multicolumn{1}{c}{GenAI-Bench} & \multicolumn{2}{c}{TIFA} & \multicolumn{2}{c}{RichHF} \\
          &  PLCC & PLCC & SRCC & PLCC & SRCC \\
        \midrule
        CLIPScore  &  0.164 & 0.309 & 0.300 & 0.302 & 0.057 \\
        ImageReward &  0.350 & 0.633 & 0.621 & 0.329 & 0.274 \\
        PickScore  &  0.354 & 0.413 & 0.392 & 0.346 & 0.340 \\
        HPSv2 &  0.139 & 0.380 & 0.365 & 0.258 & 0.187 \\
        VQAScore  &  0.499 & 0.659 & 0.695 & 0.409 & 0.483 \\
        EvalMuse & 0.498 &   0.712 & 0.749 & 0.549 & 0.518      \\
        HPSv3  &  0.139 & 0.485 & 0.484 & 0.205 & 0.184 \\
        \midrule
        \textbf{ImageDoctor} &  \textbf{0.514} & \textbf{0.740} & \textbf{0.764} & \textbf{0.808} & \textbf{0.799}\\
        \bottomrule
    \end{tabular}}
    \label{tab:cross_domain}
    \end{minipage}
    \hfill
    \begin{minipage}[c]{0.58\textwidth}
    \centering
    \vspace{0pt}
    \caption{\textbf{Ablation study} of the proposed modules. } 
    \resizebox{\textwidth}{!}{
    \begin{tabular}{l| c c | c c c c}
        \toprule
        \multirow{2}{*}{\textbf{Settings}}  & \multicolumn{2}{c}{Average} & \multicolumn{2}{|c}{Artifact} & \multicolumn{2}{c}{Misalignment} \\ 
        \cmidrule(lr){2-3} \cmidrule(lr){4-5} \cmidrule(lr){6-7} 
          & PLCC $\uparrow$ & SRCC $\uparrow$ &CC $\uparrow$ & KLD $\downarrow$  & CC $\uparrow$ & KLD $\downarrow$  \\ \midrule
        \textbf{Cold Start Stage 1}   & 0.660  & 0.656  & -  & -  & -  & -  \\ 
        + Heatmap  & 0.655  & 0.650  & 0.532  & 1.597  & 0.165  &3.031  \\ 
        + Heatmap w/o task token & 0.653  & 0.645 & 0.508  & 1.728  & 0.123  & 3.231  \\ \hline
        \textbf{Cold Start Stage 2}   & 0.720  & 0.707  & 0.558  & 1.533  & 0.224  & 2.982  \\ 
        \ \ w/o \textit{``look"}  & 0.714  & 0.705  & 0.534 & 1.599  & 0.160  &3.131  \\ 
        \ \ w/o \textit{``think"}  & 0.708  & 0.698  & 0.542  & 1.592  & 0.190  &3.038  \\ \hline
        \textbf{Reinforcement Finetuning}  & 0.741  & 0.724  & 0.571  & 1.477  & 0.225  & 2.863  \\ 
         \ \ w/o grounding reward & 0.734  & 0.718  & 0.566  & 1.507  & 0.225  & 2.865  \\ \bottomrule

    \end{tabular}
    }
    \label{tab:ablation}
    \end{minipage}
\vspace{-10pt}
\end{table}

\begin{figure}[htb]
    \centering
    \includegraphics[width=0.85\linewidth]{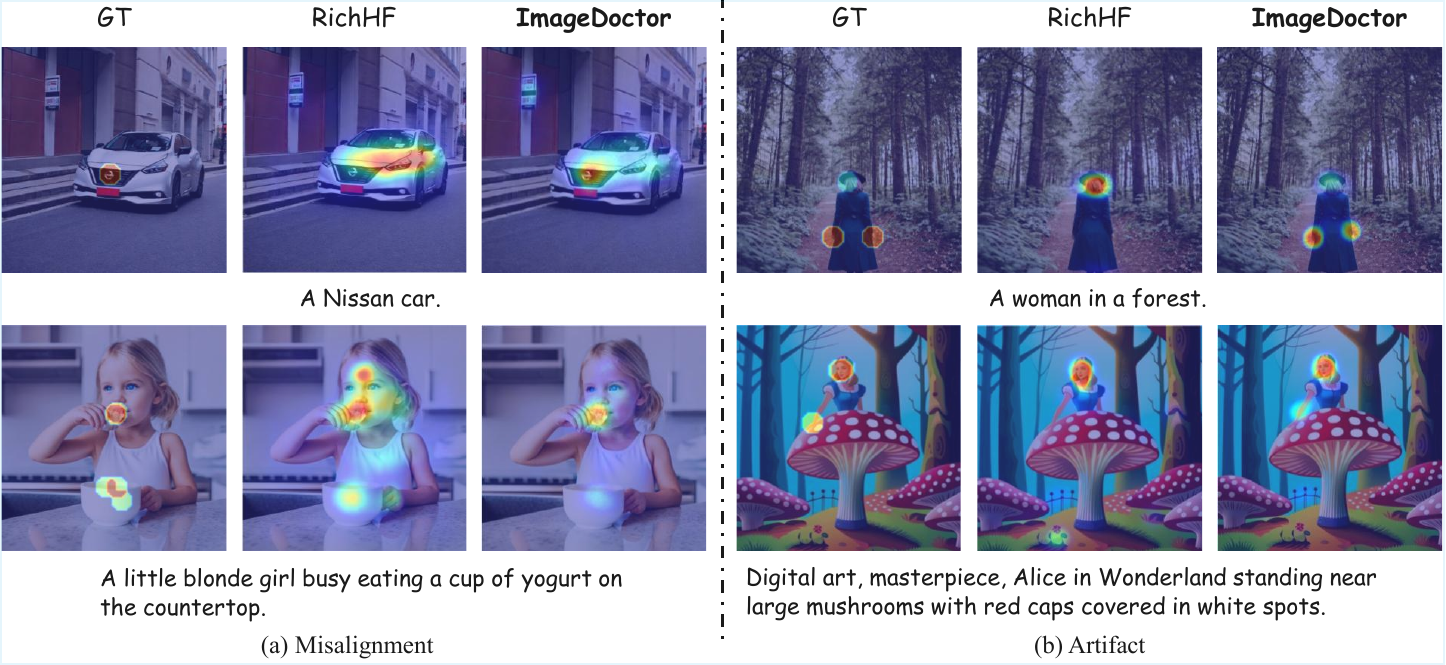}
    \caption{Visualization of misalignment and artifact heatmaps. }
    \label{fig:heatmap_vis}
    \vspace{-10pt}
\end{figure}

\noindent \textbf{Heatmap Visualization.} In~\cref{fig:heatmap_vis}, we present qualitative examples of heatmap predictions generated by ImageDoctor. For the misalignment heatmaps (\cref{fig:heatmap_vis} (a)), our model accurately localizes objects that fail to correspond to the prompt while producing fewer false positives. For the artifact heatmaps (\cref{fig:heatmap_vis} (b)), ImageDoctor effectively highlights all the regions containing artifacts, demonstrating precise spatial grounding of visual flaws. More examples are provided in~\cref{sec:add_heatmaps}.

\vspace{-6pt}

\subsection{Ablation Study}
\label{subsub:ablation}

We conduct ablation experiments on the RichHF-18K dataset to analyze the contribution of each proposed module in ImageDoctor. The results are shown in~\Cref{tab:ablation}.

\looseness -1 \noindent \textbf{Task token for heatmap decoder.} We introduce the special task tokens in the MLLM backbone to guide the heatmap prediction in the heatmap decoder. To demonstrate its effectiveness, we finetune the ImageDoctor model after cold start stage 1 on both score and heatmap prediction tasks with and without the task tokens. As shown in~\Cref{tab:ablation} (rows 2 and 3), removing the task tokens results in notable decrease of heatmap prediction performance, \eg, CC drops by 0.024 and 0.042 for artifact and misalignment, respectively.

\looseness -1 \noindent \textbf{Effect of \textit{``look"} and \textit{``think"}.} 
We propose a \textit{``look-think-prediction"} paradigm for T2I evaluation, where the model first localizes potential flawed regions by predicting bounding boxes (\textit{\textbf{``look"}}), and then analyzes and reasons about these flaws (\textit{\textbf{``think"}}) before making predictions. We conduct ablation studies assessing their contribution in~\Cref{tab:ablation} (rows 5 and 6). Removing either component leads to a performance drop. In particular, \textit{``think"} plays a more critical role in score accuracy, with PLCC decreasing from 0.720 to 0.708 without \textit{``think"}, while \textit{``look"} provides stronger benefits for heatmap prediction, where misalignment CC falls from 0.224 to 0.160 without \textit{``look"}. These results highlight the complementary roles of \textit{``look"} and \textit{``think"}: the former enhances spatial localization of flaws, while the latter strengthens semantic reasoning for accurate evaluation.

\looseness -1 \noindent \textbf{Effect of grounding reward.} 
We introduce a grounding reward in reinforcement finetuning to encourage the model to accurately localize flawed regions. As shown in~\Cref{tab:ablation} (row 8), removing the grounding reward leads to a decline in score prediction performance. Moreover, it also causes a notable drop in artifact heatmap quality. These results hightlight the importance of grounding reward.  

\vspace{-6pt}

\section{Application in Downstream Tasks}

To further demonstrate the effectiveness of ImageDoctor’s rich feedback, we explore its application in downstream tasks, specifically as a verifier in test-time scaling and a reward function for reinforcement learning of T2I model.

\subsection{ImageDoctor as a Verifier for Test-time Scaling}

\begin{wrapfigure}{r}{0.55\textwidth}
    \centering
    \vspace{-12pt}
    \includegraphics[width=1\linewidth]{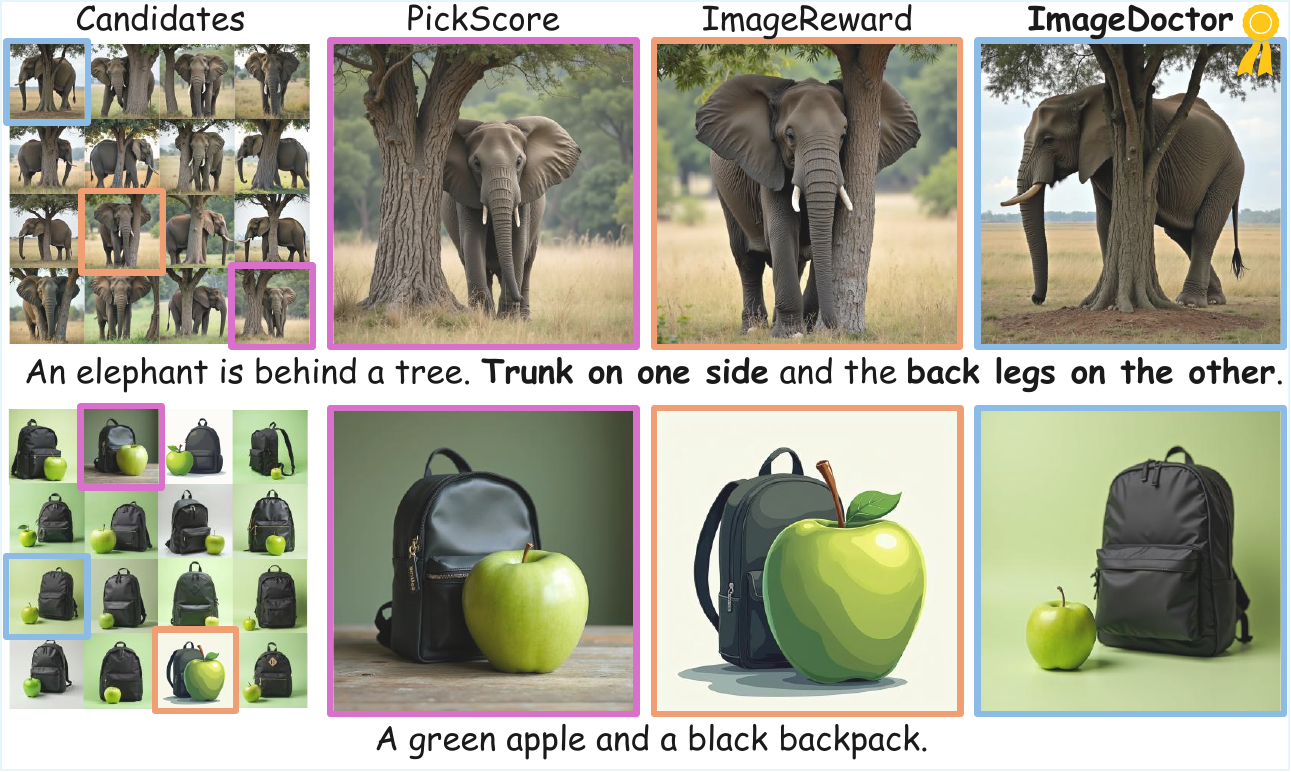}
    \caption{Qualitative comparison on selected images by different verifiers in test-time scaling. ImageDoctor picks the images that faithfully reflect the text prompt (\textit{top}) and preserve realistic object scale (\textit{bottom}). }
    \label{fig:verifier_vis}
    \vspace{-10pt}
\end{wrapfigure}

Recent works have explored test-time scaling~\citep{ma2025inference, guo2025can} for improving diffusion model performance by generating multiple samples during inference and searching for the best candidate. This approach requires a verifier to distinguish subtle differences among generated images and reliably select the best candidate. We test ImageDoctor as an image verifier, where we sample 16 images for a given prompt and select the best candidates leveraging the four-dimensional scores. The images are generated at a resolution of $1024 \times 1024$ using the Flux-dev~\citep{flux2024} model. Visualization results in~\cref{fig:verifier_vis} show that ImageDoctor reliably selects images that better align with the prompt, often preferring those with more realistic and coherent details.

\subsection{ImageDoctor as a Reward function}

\looseness -1 \noindent 
\textbf{Setup.} We use Stable Diffusion 3.5-medium as the base model and demonstrate the results of using ImageDoctor as a reward function in Flow-GRPO as well as the proposed DenseFlow-GRPO (Sec.~\ref{sec:denseflow-grpo}). We train the models for 1,300 iterations on the Pick-a-Pic prompts~\citep{kirstain2023pick}. We evaluate the base and finetuned model performance on DrawBench~\citep{saharia2022photorealistic} using ImageReward~\citep{xu2023imagereward}, CLIPScore~\citep{hessel2021clipscore} and UnifiedReward~\citep{unifiedreward} as the metrics.

\looseness -1 \noindent \textbf{Results with Flow-GRPO.} Flow-GRPO adopts score-only reward functions for training T2I model. We use ImageDoctor predicted scores as the reward function, and compare the performance of using PickScore~\citep{kirstain2023pick} and RichHF~\citep{liang2024rich}. As shown in~\Cref{tab:reward}, using ImageDoctor as the reward function consistently offers the highest gain across all evaluation metrics thanks to its strong ability in predicting accurate image evaluation scores.

\begin{center}
\vspace{-10pt}

\begin{minipage}[c]{0.56\textwidth}
  \vspace{0pt}\centering
  \captionof{table}{Performance on human preference scores when ImageDoctor serves as a reward function.}
  \label{tab:reward}
  \resizebox{\linewidth}{!}{%
    \begin{tabular}{c|ccc}
      \toprule
      \textbf{Reward} & ImageReward & CLIPScore & UnifiedReward \\
      \midrule
      Base & 0.818 & 0.951 & 2.903 \\
      \midrule
      \multicolumn{4}{c}{\textbf{Flow-GRPO}} \\
      \midrule
      PickScore & 1.002 & 0.941 & 2.940 \\
      RichHF    & 0.879 & 0.944 & 2.921 \\
      \textbf{ImageDoctor} & 1.029 & 0.956 & 2.960 \\
      \midrule
      \multicolumn{4}{c}{\textbf{DenseFlow-GRPO}} \\
      \midrule
      \textbf{ImageDoctor} & 1.100 & 0.969 & 3.000 \\
      \bottomrule
    \end{tabular}%
  }
\end{minipage}\hfill
\begin{minipage}[c]{0.4\textwidth}
  \vspace{0pt}\centering
  \includegraphics[width=\linewidth]{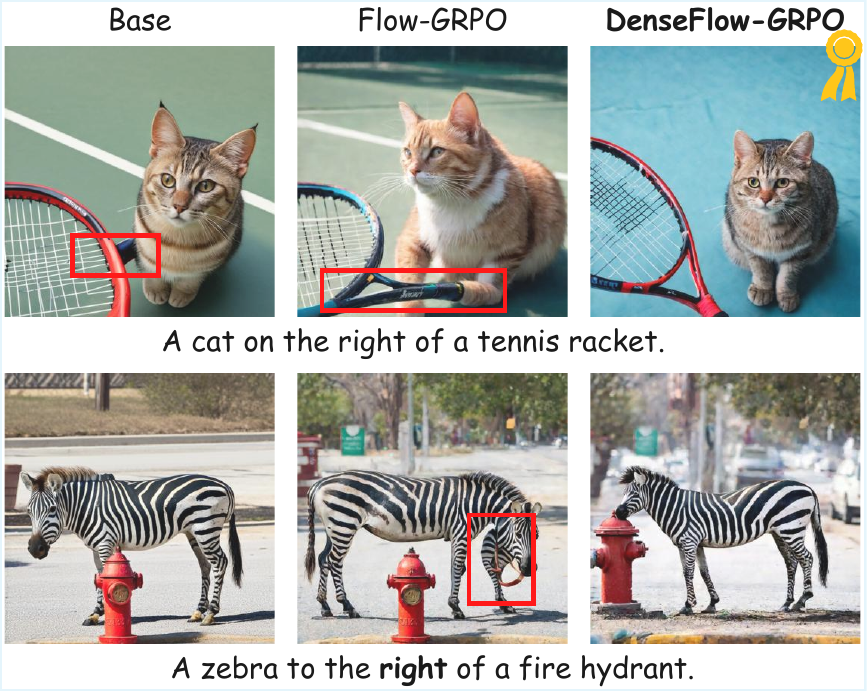}
  \captionof{figure}{Flow-GRPO vs. DenseFlow-GRPO. The artifacts are \textcolor{red}{{boxed}}.}
  \label{fig:reward_vis}
\end{minipage}
\vspace{-6pt}
\end{center}

\looseness -1 \noindent \textbf{Results with DenseFlow-GRPO.} ImageDoctor is capable of generating spatial heatmaps and scalar scores, making it well-suited for pixel-level feedback. To leverage this property, we introduce DenseFlow-GRPO, which incorporates heatmap-guided dense rewards for more fine-grained optimization. As shown in~\Cref{tab:reward}, by combining scalar scores with heatmaps, DenseFlow-GRPO achieves the best overall results and outperforms the Flow-GRPO variant with ImageDoctor score prediction as the reward function. These findings demonstrate that ImageDoctor’s dense, multi-aspect feedback provides fine-grained supervision and leads to consistently stronger alignment with human preference. We provide a visual comparison of Flow-GRPO and DenseFlow-GRPO in~\cref{fig:reward_vis}. Flow-GRPO adopts an image-level formulation that is often insufficient for removing localized artifacts, as the reward signal provides a sparse global score for the entire image, while DenseGRPO's heatmap-based dense reward design can target and refine local details to eliminate such flaws.
\vspace{-8pt}

\section{Conclusion}
\vspace{-6pt}

In this work, we introduce ImageDoctor, a unified evaluation framework for text-to-image generation that produces both multi-aspect scores and spatially grounded heatmaps. To enhance the evaluation accuracy and interpretability, we propose a \textit{``look-think-predict"} paradigm, which localizes flaws, analyzes them, and delivers a final judgment. Furthermore, we propose DenseFlow-GRPO that utilizes the dense rewards generated by ImageDoctor for finetuning T2I model. Extensive experiments demonstrate its versatility—serving as a metric, verifier, and reward function—showing that ImageDoctor provides robust, interpretable, and human-aligned feedback for generated images.

\bibliography{iclr2026_conference}

\begin{thebibliography}{39}
\providecommand{\natexlab}[1]{#1}
\providecommand{\url}[1]{\texttt{#1}}
\expandafter\ifx\csname urlstyle\endcsname\relax
  \providecommand{\doi}[1]{doi: #1}\else
  \providecommand{\doi}{doi: \begingroup \urlstyle{rm}\Url}\fi

\bibitem[Bai et~al.(2025)Bai, Chen, Liu, Wang, Ge, Song, Dang, Wang, Wang, Tang, Zhong, Zhu, Yang, Li, Wan, Wang, Ding, Fu, Xu, Ye, Zhang, Xie, Cheng, Zhang, Yang, Xu, and Lin]{Qwen2.5-VL}
Shuai Bai, Keqin Chen, Xuejing Liu, Jialin Wang, Wenbin Ge, Sibo Song, Kai Dang, Peng Wang, Shijie Wang, Jun Tang, Humen Zhong, Yuanzhi Zhu, Mingkun Yang, Zhaohai Li, Jianqiang Wan, Pengfei Wang, Wei Ding, Zheren Fu, Yiheng Xu, Jiabo Ye, Xi~Zhang, Tianbao Xie, Zesen Cheng, Hang Zhang, Zhibo Yang, Haiyang Xu, and Junyang Lin.
\newblock Qwen2.5-vl technical report.
\newblock \emph{arXiv preprint arXiv:2502.13923}, 2025.

\bibitem[Comanici et~al.(2025)Comanici, Bieber, Schaekermann, Pasupat, Sachdeva, Dhillon, Blistein, Ram, Zhang, Rosen, et~al.]{comanici2025gemini}
Gheorghe Comanici, Eric Bieber, Mike Schaekermann, Ice Pasupat, Noveen Sachdeva, Inderjit Dhillon, Marcel Blistein, Ori Ram, Dan Zhang, Evan Rosen, et~al.
\newblock Gemini 2.5: Pushing the frontier with advanced reasoning, multimodality, long context, and next generation agentic capabilities.
\newblock \emph{arXiv preprint arXiv:2507.06261}, 2025.

\bibitem[Croitoru et~al.(2023)Croitoru, Hondru, Ionescu, and Shah]{croitoru2023diffusion}
Florinel-Alin Croitoru, Vlad Hondru, Radu~Tudor Ionescu, and Mubarak Shah.
\newblock Diffusion models in vision: A survey.
\newblock \emph{PAMI}, 2023.

\bibitem[Esser et~al.(2024)Esser, Kulal, Blattmann, Entezari, M{\"u}ller, Saini, Levi, Lorenz, Sauer, Boesel, et~al.]{esser2024scaling}
Patrick Esser, Sumith Kulal, Andreas Blattmann, Rahim Entezari, Jonas M{\"u}ller, Harry Saini, Yam Levi, Dominik Lorenz, Axel Sauer, Frederic Boesel, et~al.
\newblock Scaling rectified flow transformers for high-resolution image synthesis.
\newblock In \emph{ICML}, 2024.

\bibitem[Goodfellow et~al.(2014)Goodfellow, Pouget-Abadie, Mirza, Xu, Warde-Farley, Ozair, Courville, and Bengio]{goodfellow2014generative}
Ian~J Goodfellow, Jean Pouget-Abadie, Mehdi Mirza, Bing Xu, David Warde-Farley, Sherjil Ozair, Aaron Courville, and Yoshua Bengio.
\newblock Generative adversarial nets.
\newblock In \emph{NeurIPS}, 2014.

\bibitem[Gu et~al.(2023)Gu, Zhai, Zhang, Susskind, and Jaitly]{gu2023matryoshka}
Jiatao Gu, Shuangfei Zhai, Yizhe Zhang, Joshua~M Susskind, and Navdeep Jaitly.
\newblock Matryoshka diffusion models.
\newblock In \emph{ICLR}, 2023.

\bibitem[Guo et~al.(2025)Guo, Zhang, Tong, Zhao, Gao, Li, and Heng]{guo2025can}
Ziyu Guo, Renrui Zhang, Chengzhuo Tong, Zhizheng Zhao, Peng Gao, Hongsheng Li, and Pheng-Ann Heng.
\newblock Can we generate images with cot? let's verify and reinforce image generation step by step.
\newblock \emph{arXiv preprint arXiv:2501.13926}, 2025.

\bibitem[He et~al.(2016)He, Zhang, Ren, and Sun]{he2016deep}
Kaiming He, Xiangyu Zhang, Shaoqing Ren, and Jian Sun.
\newblock Deep residual learning for image recognition.
\newblock In \emph{{CVPR}}, 2016.

\bibitem[Hessel et~al.(2021)Hessel, Holtzman, Forbes, Le~Bras, and Choi]{hessel2021clipscore}
Jack Hessel, Ari Holtzman, Maxwell Forbes, Ronan Le~Bras, and Yejin Choi.
\newblock Clipscore: A reference-free evaluation metric for image captioning.
\newblock In \emph{EMNLP}, 2021.

\bibitem[Ho et~al.(2020)Ho, Jain, and Abbeel]{ho2020denoising}
Jonathan Ho, Ajay Jain, and Pieter Abbeel.
\newblock Denoising diffusion probabilistic models.
\newblock In \emph{NeurIPS}, 2020.

\bibitem[Hu et~al.(2023)Hu, Liu, Kasai, Wang, Ostendorf, Krishna, and Smith]{hu2023tifa}
Yushi Hu, Benlin Liu, Jungo Kasai, Yizhong Wang, Mari Ostendorf, Ranjay Krishna, and Noah~A Smith.
\newblock Tifa: Accurate and interpretable text-to-image faithfulness evaluation with question answering.
\newblock In \emph{{CVPR}}, 2023.

\bibitem[Kingma \& Welling(2013)Kingma and Welling]{kingma2013auto}
Diederik~P Kingma and Max Welling.
\newblock Auto-encoding variational bayes.
\newblock \emph{arXiv preprint arXiv:1312.6114}, 2013.

\bibitem[Kirillov et~al.(2023)Kirillov, Mintun, Ravi, Mao, Rolland, Gustafson, Xiao, Whitehead, Berg, Lo, et~al.]{kirillov2023segment}
Alexander Kirillov, Eric Mintun, Nikhila Ravi, Hanzi Mao, Chloe Rolland, Laura Gustafson, Tete Xiao, Spencer Whitehead, Alexander~C Berg, Wan-Yen Lo, et~al.
\newblock Segment anything.
\newblock In \emph{CVPR}, 2023.

\bibitem[Kirstain et~al.(2023)Kirstain, Polyak, Singer, Matiana, Penna, and Levy]{kirstain2023pick}
Yuval Kirstain, Adam Polyak, Uriel Singer, Shahbuland Matiana, Joe Penna, and Omer Levy.
\newblock {Pick-a-Pic: An open dataset of user preferences for text-to-image generation}.
\newblock In \emph{{NeurIPS}}, 2023.

\bibitem[Labs(2024)]{flux2024}
Black~Forest Labs.
\newblock Flux.
\newblock \url{https://github.com/black-forest-labs/flux}, 2024.

\bibitem[Lee et~al.(2023)Lee, Yasunaga, Meng, Mai, Park, Gupta, Zhang, Narayanan, Teufel, Bellagente, et~al.]{lee2023holistic}
Tony Lee, Michihiro Yasunaga, Chenlin Meng, Yifan Mai, Joon~Sung Park, Agrim Gupta, Yunzhi Zhang, Deepak Narayanan, Hannah Teufel, Marco Bellagente, et~al.
\newblock Holistic evaluation of text-to-image models.
\newblock In \emph{NeurIPS}, 2023.

\bibitem[Li et~al.(2024)Li, Lin, Pathak, Li, Fei, Wu, Xia, Zhang, Neubig, and Ramanan]{li2024evaluating}
Baiqi Li, Zhiqiu Lin, Deepak Pathak, Jiayao Li, Yixin Fei, Kewen Wu, Xide Xia, Pengchuan Zhang, Graham Neubig, and Deva Ramanan.
\newblock Evaluating and improving compositional text-to-visual generation.
\newblock In \emph{CVPR}, 2024.

\bibitem[Li et~al.(2022)Li, Li, Xiong, and Hoi]{li2022blip}
Junnan Li, Dongxu Li, Caiming Xiong, and Steven Hoi.
\newblock Blip: Bootstrapping language-image pre-training for unified vision-language understanding and generation.
\newblock In \emph{ICML}, 2022.

\bibitem[Liang et~al.(2024)Liang, He, Li, Li, Klimovskiy, Carolan, Sun, Pont-Tuset, Young, Yang, et~al.]{liang2024rich}
Youwei Liang, Junfeng He, Gang Li, Peizhao Li, Arseniy Klimovskiy, Nicholas Carolan, Jiao Sun, Jordi Pont-Tuset, Sarah Young, Feng Yang, et~al.
\newblock Rich human feedback for text-to-image generation.
\newblock In \emph{{CVPR}}, 2024.

\bibitem[Liu et~al.(2025)Liu, Liu, Liang, Li, Liu, Wang, Wan, Zhang, and Ouyang]{liu2025flow}
Jie Liu, Gongye Liu, Jiajun Liang, Yangguang Li, Jiaheng Liu, Xintao Wang, Pengfei Wan, Di~Zhang, and Wanli Ouyang.
\newblock Flow-grpo: Training flow matching models via online rl.
\newblock \emph{arXiv preprint arXiv:2505.05470}, 2025.

\bibitem[Ma et~al.(2025{\natexlab{a}})Ma, Tong, Jia, Hu, Su, Zhang, Yang, Li, Jaakkola, Jia, and Xie]{ma2025inference}
Nanye Ma, Shangyuan Tong, Haolin Jia, Hexiang Hu, Yu-Chuan Su, Mingda Zhang, Xuan Yang, Yandong Li, Tommi Jaakkola, Xuhui Jia, and Saining Xie.
\newblock Scaling inference time compute for diffusion models.
\newblock In \emph{CVPR}, 2025{\natexlab{a}}.

\bibitem[Ma et~al.(2025{\natexlab{b}})Ma, Wu, Sun, and Li]{ma2025hpsv3}
Yuhang Ma, Xiaoshi Wu, Keqiang Sun, and Hongsheng Li.
\newblock Hpsv3: Towards wide-spectrum human preference score.
\newblock \emph{arXiv preprint arXiv:2508.03789}, 2025{\natexlab{b}}.

\bibitem[Nichol et~al.(2021)Nichol, Dhariwal, Ramesh, Shyam, Mishkin, McGrew, Sutskever, and Chen]{nichol2021glide}
Alex Nichol, Prafulla Dhariwal, Aditya Ramesh, Pranav Shyam, Pamela Mishkin, Bob McGrew, Ilya Sutskever, and Mark Chen.
\newblock Glide: Towards photorealistic image generation and editing with text-guided diffusion models.
\newblock \emph{arXiv preprint arXiv:2112.10741}, 2021.

\bibitem[Radford et~al.(2021)Radford, Kim, Hallacy, Ramesh, Goh, Agarwal, Sastry, Askell, Mishkin, Clark, et~al.]{radford2021learning}
Alec Radford, Jong~Wook Kim, Chris Hallacy, Aditya Ramesh, Gabriel Goh, Sandhini Agarwal, Girish Sastry, Amanda Askell, Pamela Mishkin, Jack Clark, et~al.
\newblock Learning transferable visual models from natural language supervision.
\newblock In \emph{{ICML}}, 2021.

\bibitem[Rombach et~al.(2022)Rombach, Blattmann, Lorenz, Esser, and Ommer]{rombach2022high}
Robin Rombach, Andreas Blattmann, Dominik Lorenz, Patrick Esser, and Bj{\"o}rn Ommer.
\newblock High-resolution image synthesis with latent diffusion models.
\newblock In \emph{CVPR}, 2022.

\bibitem[Saharia et~al.(2022)Saharia, Chan, Saxena, Li, Whang, Denton, Ghasemipour, Gontijo~Lopes, Karagol~Ayan, Salimans, et~al.]{saharia2022photorealistic}
Chitwan Saharia, William Chan, Saurabh Saxena, Lala Li, Jay Whang, Emily~L Denton, Kamyar Ghasemipour, Raphael Gontijo~Lopes, Burcu Karagol~Ayan, Tim Salimans, et~al.
\newblock Photorealistic text-to-image diffusion models with deep language understanding.
\newblock In \emph{NeurIPS}, 2022.

\bibitem[Shao et~al.(2024)Shao, Wang, Zhu, Xu, Song, Bi, Zhang, Zhang, Li, Wu, et~al.]{shao2024deepseekmath}
Zhihong Shao, Peiyi Wang, Qihao Zhu, Runxin Xu, Junxiao Song, Xiao Bi, Haowei Zhang, Mingchuan Zhang, YK~Li, Yang Wu, et~al.
\newblock Deepseekmath: Pushing the limits of mathematical reasoning in open language models.
\newblock \emph{arXiv preprint arXiv:2402.03300}, 2024.

\bibitem[Simonyan et~al.(2013)Simonyan, Vedaldi, and Zisserman]{simonyan2013deep}
Karen Simonyan, Andrea Vedaldi, and Andrew Zisserman.
\newblock Deep inside convolutional networks: Visualising image classification models and saliency maps.
\newblock \emph{arXiv preprint arXiv:1312.6034}, 2013.

\bibitem[Tian et~al.(2024)Tian, Jiang, Yuan, Peng, and Wang]{tian2024visual}
Keyu Tian, Yi~Jiang, Zehuan Yuan, Bingyue Peng, and Liwei Wang.
\newblock Visual autoregressive modeling: Scalable image generation via next-scale prediction.
\newblock In \emph{NeurIPS}, 2024.

\bibitem[Wang et~al.(2025{\natexlab{a}})Wang, Li, Zang, Wang, Lu, Jin, and Wang]{unifiedreward-think}
Yibin Wang, Zhimin Li, Yuhang Zang, Chunyu Wang, Qinglin Lu, Cheng Jin, and Jiaqi Wang.
\newblock Unified multimodal chain-of-thought reward model through reinforcement fine-tuning.
\newblock \emph{arXiv preprint arXiv:2505.03318}, 2025{\natexlab{a}}.

\bibitem[Wang et~al.(2025{\natexlab{b}})Wang, Zang, Li, Jin, and Wang]{unifiedreward}
Yibin Wang, Yuhang Zang, Hao Li, Cheng Jin, and Jiaqi Wang.
\newblock Unified reward model for multimodal understanding and generation.
\newblock \emph{arXiv preprint arXiv:2503.05236}, 2025{\natexlab{b}}.

\bibitem[Wang et~al.(2025{\natexlab{c}})Wang, Chen, Hu, Liu, Sun, Wu, Su, Yu, Barsoum, and Liu]{wang2025instella}
Ze~Wang, Hao Chen, Benran Hu, Jiang Liu, Ximeng Sun, Jialian Wu, Yusheng Su, Xiaodong Yu, Emad Barsoum, and Zicheng Liu.
\newblock Instella-t2i: Pushing the limits of 1d discrete latent space image generation.
\newblock \emph{arXiv preprint arXiv:2506.21022}, 2025{\natexlab{c}}.

\bibitem[Wu et~al.(2023{\natexlab{a}})Wu, Hao, Sun, Chen, Zhu, Zhao, and Li]{wu2023human2}
Xiaoshi Wu, Yiming Hao, Keqiang Sun, Yixiong Chen, Feng Zhu, Rui Zhao, and Hongsheng Li.
\newblock Human preference score v2: A solid benchmark for evaluating human preferences of text-to-image synthesis.
\newblock \emph{arXiv preprint arXiv:2306.09341}, 2023{\natexlab{a}}.

\bibitem[Wu et~al.(2023{\natexlab{b}})Wu, Sun, Zhu, Zhao, and Li]{wu2023human1}
Xiaoshi Wu, Keqiang Sun, Feng Zhu, Rui Zhao, and Hongsheng Li.
\newblock Human preference score: Better aligning text-to-image models with human preference.
\newblock In \emph{CVPR}, 2023{\natexlab{b}}.

\bibitem[Xie et~al.(2024)Xie, Mao, Bai, Zhang, Wang, Lin, Gu, Chen, Yang, and Shou]{xie2024show}
Jinheng Xie, Weijia Mao, Zechen Bai, David~Junhao Zhang, Weihao Wang, Kevin~Qinghong Lin, Yuchao Gu, Zhijie Chen, Zhenheng Yang, and Mike~Zheng Shou.
\newblock Show-o: One single transformer to unify multimodal understanding and generation.
\newblock \emph{arXiv preprint arXiv:2408.12528}, 2024.

\bibitem[Xu et~al.(2023)Xu, Liu, Wu, Tong, Li, Ding, Tang, and Dong]{xu2023imagereward}
Jiazheng Xu, Xiao Liu, Yuchen Wu, Yuxuan Tong, Qinkai Li, Ming Ding, Jie Tang, and Yuxiao Dong.
\newblock Imagereward: Learning and evaluating human preferences for text-to-image generation.
\newblock In \emph{{NeurIPS}}, 2023.

\bibitem[Xue et~al.(2025)Xue, Wu, Gao, Kong, Zhu, Chen, Liu, Liu, Guo, Huang, et~al.]{xue2025dancegrpo}
Zeyue Xue, Jie Wu, Yu~Gao, Fangyuan Kong, Lingting Zhu, Mengzhao Chen, Zhiheng Liu, Wei Liu, Qiushan Guo, Weilin Huang, et~al.
\newblock Dancegrpo: Unleashing grpo on visual generation.
\newblock \emph{arXiv preprint arXiv:2505.07818}, 2025.

\bibitem[Zhao et~al.(2024)Zhao, Shi, Yu, Zhou, and Lu]{zhao2024flowturbo}
Wenliang Zhao, Minglei Shi, Xumin Yu, Jie Zhou, and Jiwen Lu.
\newblock Flowturbo: Towards real-time flow-based image generation with velocity refiner.
\newblock In \emph{NeurIPS}, 2024.

\bibitem[Zheng et~al.(2025)Zheng, Liu, Li, Chen, Yu, Gao, Dang, Liu, Men, Yang, et~al.]{zheng2025group}
Chujie Zheng, Shixuan Liu, Mingze Li, Xiong-Hui Chen, Bowen Yu, Chang Gao, Kai Dang, Yuqiong Liu, Rui Men, An~Yang, et~al.
\newblock Group sequence policy optimization.
\newblock \emph{arXiv preprint arXiv:2507.18071}, 2025.

\end{thebibliography}
\bibliographystyle{iclr2026_conference}
\clearpage
\appendix

\section*{Appendix}

\section{Details of Heatmap Decoder}
\label{sec:decoder}
To equip the MLLM with the ability to generate accurate heatmaps, we design a lightweight heatmap decoder. The decoder takes image features extracted by the visual encoder, along with a learned heatmap token and a task token that specifies the type of heatmap to be produced. 
First, the special tokens (task and heatmap) serve as queries to attend over image embedding, after which a multi-layer perceptron (MLP) updates all special tokens. Next, the updated special tokens are used as keys and values to and the image embeddings act as queries to refine the image features. The updated image embeddings are then passed through a series of convolution and deconvolution layers to upscale to the original image size. Finally, before applying the sigmoid activation, we introduce an additional token-to-image attention: the updated image embeddings attend once more to the special token embeddings. The attended heatmap token features are projected through another MLPs, and their outputs are combined with the upsampled image embeddings via a spatial point-wise product. This design strengthens the role of the task tokens in guiding the final heatmap prediction, ensuring that both semantic reasoning and localized visual evidence contribute to the spatial diagnosis. 

\begin{figure}[h]
    \centering
    \includegraphics[width=0.6\linewidth]{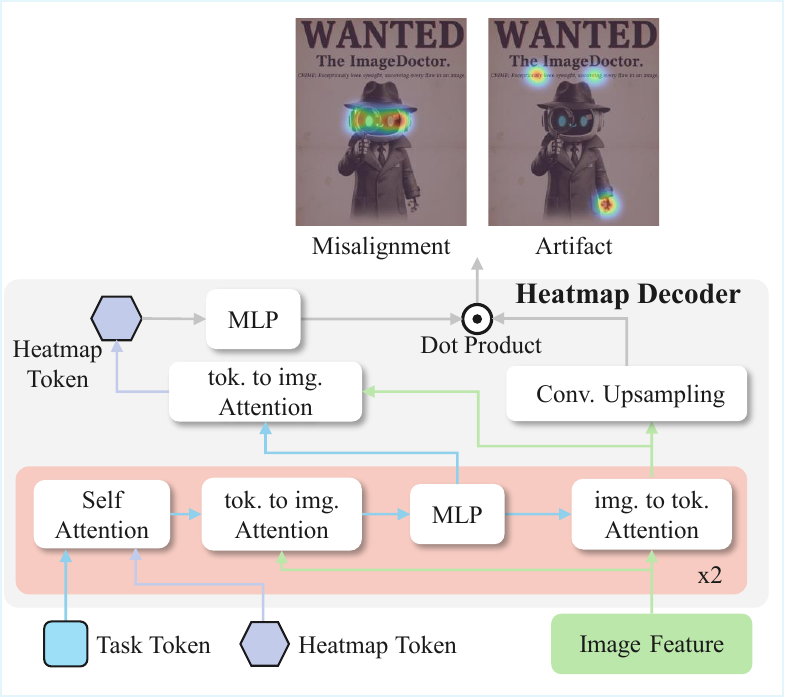}
    \caption{The architecture of heatmap decoder.}
    \label{fig:decoder}
\end{figure}

\section{Grounding Image Reasoning Generation}
\label{sec:reasoning_generation}

For reasoning path generation in Stage 2 of the cold start phase, we employ Gemini-flash with carefully designed prompts to produce detailed reasoning that bridges the image and its human annotations. Since advanced VLMs possess strong visual grounding capabilities, we enrich the input beyond the original image–prompt pair by also providing human-annotated heatmaps indicating the locations of artifacts and misalignments. To preserve the fidelity of the original image, we highlight these flawed regions using circled outlines derived from the heatmaps, rather than directly overlaying them, which enables the VLM to more accurately localize unsatisfactory regions. By combining these localized observations with human-provided scores, Gemini 2.5 Flash generates reasoning that is more precise and relevant to the evaluated image. This high-quality reasoning chain is then used to fine-tune Qwen2.5-VL, equipping it with structured and grounded evaluation capabilities.

\begin{tcolorbox}[breakable,colback=orange!5!white, colframe=orange!80!black, title=Prompt for Gemini 2.5 Flash 2.5 for Reasoning Data Generation]
\texttt{You are a multi-modal AI assistant tasked with generating a reasoning process for a human evaluation of a generated image. I am providing three images in a specific order:\\
1.  \textbf{The First Image provided is the 'Original Image'}: This is the image generated by a text-to-image model based on the input prompt "PROMPT". \textbf{This Original Image is the SOLE subject of your evaluation.} \\
2.  \textbf{The Second Image provided is the 'Artifact Heatmap Image'}: This image is a visual aid. It is the Original Image with an overlay that \textbf{ONLY serves to visually pinpoint the artifact locations. Its ONLY purpose is to help you locate the specified coordinates on the Original Image and then describe the visual characteristics of the artifact locations *on the Original Image*.} \\
3.  \textbf{The Third Image provided is the 'Misalignment Heatmap Image'}: This image is a visual aid. It is the Original Image with an overlay that \textbf{ONLY serves to visually pinpoint the misalignment locations. Its ONLY purpose is to help you locate the specified coordinates on the Original Image and then describe the visual characteristics of the misalignment locations *on the Original Image*}. \\
Below, you will find the human evaluation data of the Original Image for several dimensions, including scores, keyword alignment status.. Your goal is to analyze the \textbf{Original Image} and articulate a plausible step-by-step reasoning that would lead to the given scores, speaking from the perspective of the evaluator. \\
Human Evaluation Results: \\
*  \textbf{Semantic Alignment:} How well the image content corresponds to the original caption.  \\
     * Score: MISALIGNMENT SCORE \\
* \textbf{Aesthetics:} Assessment of composition, color usage, and overall artistic quality.  \\
* Score: AESTHETIC SCORE \\
* \textbf{Plausibility:} Realism and visual fidelity of the \textbf{Original Image}, including distortions or unnatural details.  \\
* Score: ARTIFACT SCORE\\
* \textbf{Overall Impression:} General subjective assessment of the image's quality.  \\
* Score: OVERALL SCORE \\
Your Task:  
    For each of the four evaluation dimensions (Semantic Alignment, Aesthetics, Plausibility, and Overall Impression), please provide a paragraph explaining your reasoning for the score, as if you were the original human evaluator assessing the \textbf{Original Image}. \\
        * Refer to specific visual elements of the \textbf{Original Image} that support your reasoning. \\ 
    * For the "Plausibility" or "Semantic Alignment" dimension:
  Refer to the \textbf{Second Image (Artifact Heatmap Image)} or \textbf{Third Image (Misalignment Heatmap Image)} to visually locate these coordinates on the \textbf{Original Image}, specifically connect your reasoning to these coordinates with the help of provided Artifact or Misalignment Heatmap Image. \textbf{Then, describe the visual nature of the artifact or misalignment locations as it appears *on the Original Image*.}\\
  * For other dimensions, if relevant, explain any potential reasons for a score less than perfect by examining the image. Consider all provided human evaluation results, including any labels or listed misaligned keywords, in your reasoning.
    Output Format for Each Dimension:
    Conclude each paragraph with a sentence in the following strict format: "Therefore, I give it a score of X.XX." \\
    \textbf{Important Instructions:}\\
    - Do not mention the artifact or misalignment heatmap images.\\
    - Use the coordinates to focus your visual inspection, but describe only what is visible in the Original Image.\\
    - Be concise and direct in each evaluation.\\
    - Do not include specific coordinates in your reasoning, just refer to the visual characteristics at those locations.\\
    Please now provide the reasoning for each dimension, focusing your analysis and descriptions on the \textbf{First Image (the 'Original Image' of the "PROMPT")} , using the \textbf{Second Image (the 'Artifact Heatmap Image')} and \textbf{Third Image (the 'Misalignment Heatmap Image')} strictly as a visual guide to locate artifact and misalignment locations on the Original Image. Do not mention the Artifact or Misalignment Heatmap Images in your reasoning, only use them to locate coordinates visually on the Original Image. As if human evaluation results and heatmap are not available, you only have the Original Image to evaluate.\\
    Be precise, concise, and strictly refer to the Original Image in all visual descriptions. For each dimension, use two sentences: one for the reasoning and one for the score conclusion.\\
    Begin Evaluation:  
}
\end{tcolorbox}

\section{RFT Formulation}
\label{sec:GRPO}
Given a pair of input $(\mI, \mP)$, ImageDoctor as the policy model $\pi_{\theta}$, generates a group of $N$ candidate responses $\{\vo^{1},\ldots,\vo^{N}\}$. For each response $\vo^{i}$, we compute a reward score $\mathcal{R}^{i}$ using a combination of reward functions. The rewards are normalized within the group to compute the group-normalized  advantage:
\begin{equation}
A^{i} \;=\; 
\frac{ \mathcal{R}^i - \text{mean}(\{\mathcal{R}^i\}_{i=1}^N)}
     { \text{std}(\{\mathcal{R}^i\}_{i=1}^N)}.
\end{equation}

We update the policy model $\pi_{\theta}$ by maximizing the GRPO objective function:
\begin{equation}
\mathcal{J}_\text{RFT}(\theta) = \frac{1}{N} \sum_{i=1}^{N} 
\left[ 
    \min \left( w^i A^i, \; \text{clip}(w^i, 1-\epsilon, 1+\epsilon) A^i \right) 
    - \beta \KL\!\left( \pi_{\theta} \;\|\; \pi_{\text{ref}} \right) 
\right],
\end{equation}
where $w^i=\frac{\pi_\theta(\vo^i|\mI, \mP)}{\pi_\text{old}(\vo^i|\mI, \mP)}$ denotes the likelihood ratio between $\pi_\theta$ and the old policy model $\pi_\text{old}$. The clipping threshold $\epsilon$ regulates the extent to which the policy model may update in each step to stabilize training. $\beta$ controls the KL divergence regularization term that constrains $\pi_\theta$ to remain close to the reference model $\pi_{\text{ref}}$, which is the model at the start of reinforcement learning.

\section{Grounding Reward Details}
\label{sec:grounding_reward}
To fulfill the three criteria mentioned in the main paper, we design three sub-rewards. First, we compute the average intensity within each bounding box and then take the mean across all boxes. This encourages the model to identify compact regions that align with the highlighted areas, yielding higher rewards when bounding boxes accurately capture potential flaws. Second, we measure the Intersection over Union (IoU) between each pair of bounding boxes and apply a penalty for large overlaps, which discourages redundant box predictions and promotes compactness in number. Finally, we compute the ratio between the area covered by the union of all bounding boxes and the total intensity of the heatmap, ensuring that the highlighted regions are fully covered. When the heatmap is blank and no bounding boxes are predicted, we assign a reward of 1. Conversely, if a heatmap contains highlighted regions but no bounding boxes are predicted, or if bounding boxes are predicted on a blank heatmap, we assign a reward of 0.

\section{Experimental Details}

\subsection{Datasets}

\textbf{RichHF-18K:} is a subset of the Pick-a-Pic, consisting of 16K training samples, 1K validation samples, and 1K test samples. For each text–image pair, two heatmaps and four fine-grained scores are annotated by a total of 27 annotators.

\textbf{GenAI-Bench}: It contains 1,600 prompts designed to cover essential visuo-linguistic compositional reasoning skills, with prompts sourced from professional graphic designers experienced in T2I systems. More than 15,000 human ratings are collected across ten different T2I models, ensuring both diversity and difficulty. 

\textbf{TIFA}: The test set includes 800 generated images based on 160 text inputs from TIFA v1.0. These images are produced by five generative models and annotated by two independent annotators, providing additional benchmarks for evaluating generalization.

\subsection{Extended Quantitative Results}
In~\Cref{tab:richhf}, we provide additional quantitative results on the RichHF-18K dataset. Compared with the main paper, we include the self-test baseline RichHF and a variant of ImageDoctor without the reasoning step with additional heatmap metric Normalized Scanpath Saliency (NSS) and AUC-Judd. This design is motivated by scenarios where only quantitative scores are needed for efficiency, such as when serving as a reward function. To enable this, we append the input prompt with a fixed reasoning template—\texttt{<think>}…\texttt{</think>} \texttt{<answer>}—as if the reasoning chain were already completed, allowing ImageDoctor to directly output scores and heatmaps. As shown in~\Cref{tab:richhf}, this variant incurs only a minor performance drop compared to the full model, while maintaining high accuracy with significantly improved speed. For this reason, we adopt this faster variant in Flow-GRPO and DenseFlow-GRPO, where efficiency in advantage computation is critical.

\begin{table}[h]
    \setlength{\tabcolsep}{4pt}
    \caption{Score prediction and heatmap prediction results on RichHF-18K. $\downarrow$ indicates lower is better and $\uparrow$ indicates higher is better. $GT=0$ refers to empty ground truth heatmap. $GT>0$ refers to heatmaps with artifact of misalignment. There are total 69 and 144 out of 955 test samples are empty for artifact and misalignment heatmaps.}

    \begin{subtable}{\linewidth}
    \centering
    \caption{Performance comparison of score prediction.}
     \resizebox{\textwidth}{!}{
    \begin{tabular}{l|cccccccc|cc}
        \toprule
        & \multicolumn{2}{c}{Plausibility} & \multicolumn{2}{c}{Aesthetics} & \multicolumn{2}{c}{Semantic Alignment} & \multicolumn{2}{c}{Overall} & \multicolumn{2}{|c}{Average} \\
        \cmidrule(lr){2-3} \cmidrule(lr){4-5} \cmidrule(lr){6-7} \cmidrule(lr){8-9} \cmidrule(lr){10-11}
        & PLCC $\uparrow$ & SRCC $\uparrow$ & PLCC $\uparrow$ & SRCC $\uparrow$ & PLCC $\uparrow$ & SRCC $\uparrow$ & PLCC $\uparrow$ & SRCC $\uparrow$ & PLCC $\uparrow$ & SRCC $\uparrow$ \\
        \midrule
        ResNet-50~\citep{he2016deep} & 0.495 & 0.487 & 0.370 & 0.363 & 0.108 & 0.119 & 0.337 & 0.308 & 0.328 & 0.319 \\
        CLIP~\citep{radford2021learning}  & 0.390 & 0.378 & 0.357 & 0.360 & 0.398 & 0.390 & 0.353 & 0.352 &0.374 & 0.370 \\
        PickScore~\citep{kirstain2023pick} & 0.010 & 0.028 & 0.131 & 0.140 & 0.346 & 0.340 & 0.202 & 0.226 & 0.172 & 0.183 \\
        RichHF~\citep{liang2024rich} & 0.693 & 0.681 & 0.600 & 0.589 & 0.474 & 0.496 & 0.580 & 0.562 & 0.586 & 0.582\\
        RichHF\footnotemark[1]~\citep{liang2024rich} & 0.704
 & 0.694 & 0.636 & 0.618 & 0.563 & 0.602 & 0.648 & 0.634 & 0.638 & 0.637\\
        \midrule
        ImageDoctor$_{\text{no think}}$ & 0.722
 & 0.712 & 0.675 & 0.656 & 0.793 &  0.792 & 0.728 & 0.700  &0.729 & 0.715 \\
        \textbf{ImageDoctor (Ours)} & \textbf{0.727} &\textbf{ 0.711} & \textbf{0.681} & \textbf{0.662} & \textbf{0.808} &  \textbf{0.799} & \textbf{0.745} & \textbf{0.725}  &\textbf{0.741} &\textbf{ 0.724} \\
        \bottomrule
    \end{tabular}
}
    
    \label{tab:score}
      \end{subtable}

  \vspace{0.6em}

    \begin{subtable}{\linewidth}
    \centering
    \caption{Performance comparison of artifact heatmap prediction.}
    \resizebox{.8\textwidth}{!}{
  \begin{tabular}{l|c|c|ccccc} 
    \toprule
    & \multicolumn{1}{c|}{All data} & \multicolumn{1}{c|}{$GT=0$} &  \multicolumn{5}{c}{$GT>0$} \\
    \cmidrule(lr){2-8}
    &\multicolumn{1}{c|}{ MSE $\downarrow$} &\multicolumn{1}{c|}{MSE $\downarrow$} & CC $\uparrow$ & KLD $\downarrow$ & SIM $\uparrow$ & NSS $\uparrow$ & AUC-Judd $\uparrow$\\
    \midrule
    ResNet-50~\citep{he2016deep}  & 0.00996 & 0.00093 & 0.506 & 1.669 & 0.338 & 2.924 & 0.909  \\
    RichHF~\citep{liang2024rich} & 0.00920 & 0.00095 & 0.556 & 1.652 & 0.409 & \textbf{3.085} & \textbf{0.913}\\
    RichHF\footnotemark[1]~\citep{liang2024rich} & 0.00920 
 & 0.00080 & 0.545 & 1.568 & 0.375 & 1.808 & 0.893\\
    
    \midrule
    ImageDoctor$_{\text{no think}}$ & 0.00879 & 0.00091 & 0.569 & 1.483 & 0.405 &1.877& 0.903\\
   \textbf{ImageDoctor (Ours)} & \textbf{0.00891} & \textbf{0.00076} & \textbf{0.571} & \textbf{1.477} & \textbf{0.412} &1.884& 0.903\\
    \bottomrule 
  \end{tabular}}
  
  \end{subtable}

    \vspace{0.6em}

    \begin{subtable}{\linewidth}
    \centering
    \caption{Performance comparison of misalignment heatmap prediction.}
    \resizebox{.88\textwidth}{!}{
  \begin{tabular}{l|c|c|ccccc} 
    \toprule
    & \multicolumn{1}{c|}{All data} & \multicolumn{1}{c|}{$GT=0$} &  \multicolumn{5}{c}{$GT>0$} \\
    \cmidrule(lr){2-8}
    &\multicolumn{1}{c|}{ MSE $\downarrow$} &\multicolumn{1}{c|}{MSE $\downarrow$} & CC $\uparrow$ & KLD $\downarrow$ & SIM $\uparrow$ & NSS $\uparrow$ & AUC-Judd $\uparrow$\\
    \midrule

    CLIP Gradient~\citep{simonyan2013deep} & 0.00817 & 0.00551 & 0.015 & 3.844 & 0.041 & 0.143 & 0.643\\
    
    RichHF~\citep{liang2024rich} & 0.00304 & 0.00006 & 0.212 & 2.933 & 0.106 & \textbf{1.411} & \textbf{0.841}\\
    RichHF\footnotemark[1]~\citep{liang2024rich} & 0.00300
 & 0.00020 & 0.219 & 2.900 & 0.099 & 1.224 & 0.794\\
    
    \midrule
    ImageDoctor$_{\text{no think}}$ & 0.00310 & \textbf{0.00001} & 0.219 & 2.890 & \textbf{0.121} &1.230& 0.796\\
   \textbf{ImageDoctor (Ours)} & \textbf{0.00299} & 0.00003 & \textbf{0.225} & \textbf{2.863} & 0.108 &1.257& 0.801\\
    \bottomrule 
  \end{tabular}}

  \end{subtable}
  \label{tab:richhf}

\end{table}

\subsection{Qualitative Results}
\begin{figure}[htb]
    \centering
    \includegraphics[width=\linewidth]{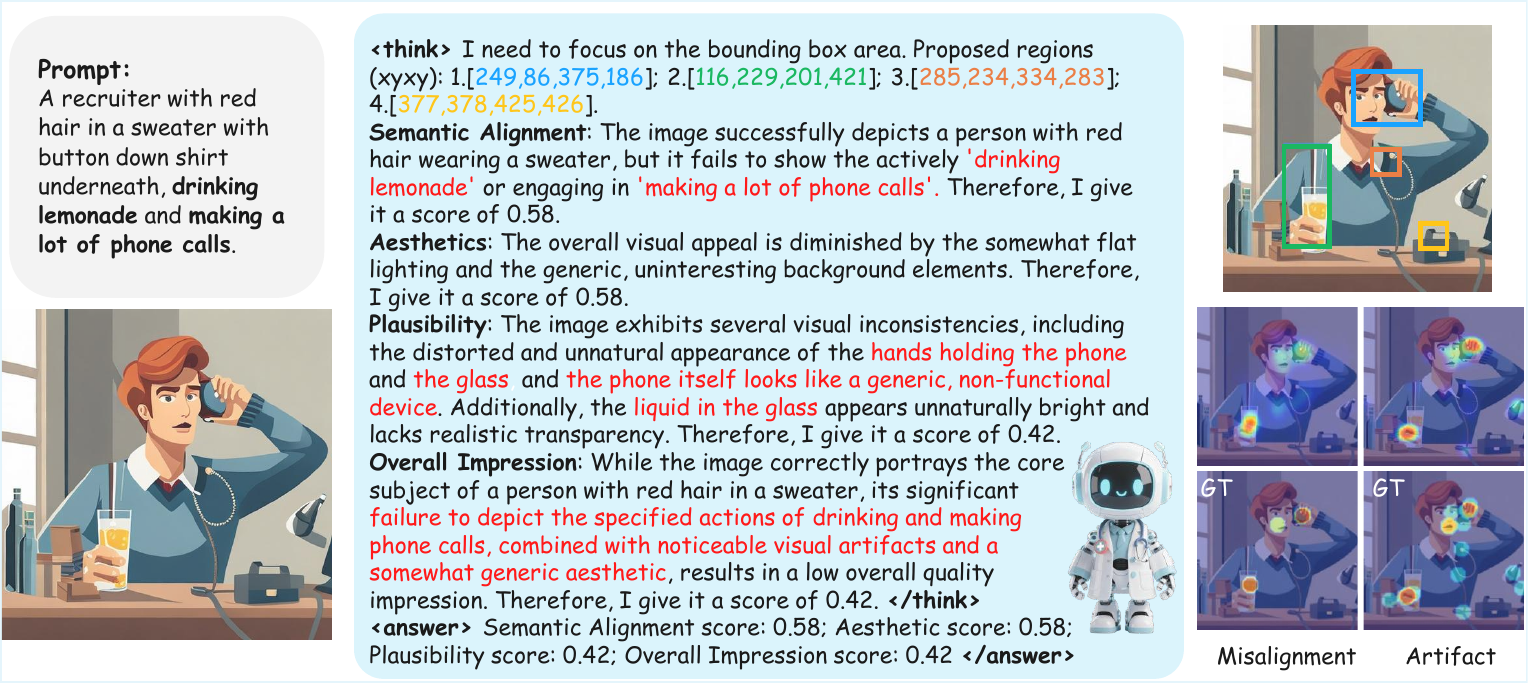}
    \caption{Qualitative Results of ImageDoctor Grounded Image Reasoning.}
    \label{fig:qualitative}
\end{figure}

\cref{fig:qualitative} presents ImageDoctor’s responses given an image–prompt pair. We observe that ImageDoctor first localizes potential flaw regions, where its reasoning and heatmap predictions closely align. For example, it correctly identifies misaligned keywords such as \textit{drinking lemonade} and \textit{making a lot of phone calls}. In addition, it detects artifacts appearing in the image, including unnatural glass shapes, distorted hands, unrealistic liquid in the glass, and the phone. Finally, the heatmaps accurately depict the misaligned and implausible areas, highlighting ImageDoctor’s strong localization and reasoning capabilities and alignment with human preferences.

\subsection{Additional Heatmap Visualization}
\label{sec:add_heatmaps}
\begin{figure}
    \centering
    \includegraphics[width=\linewidth]{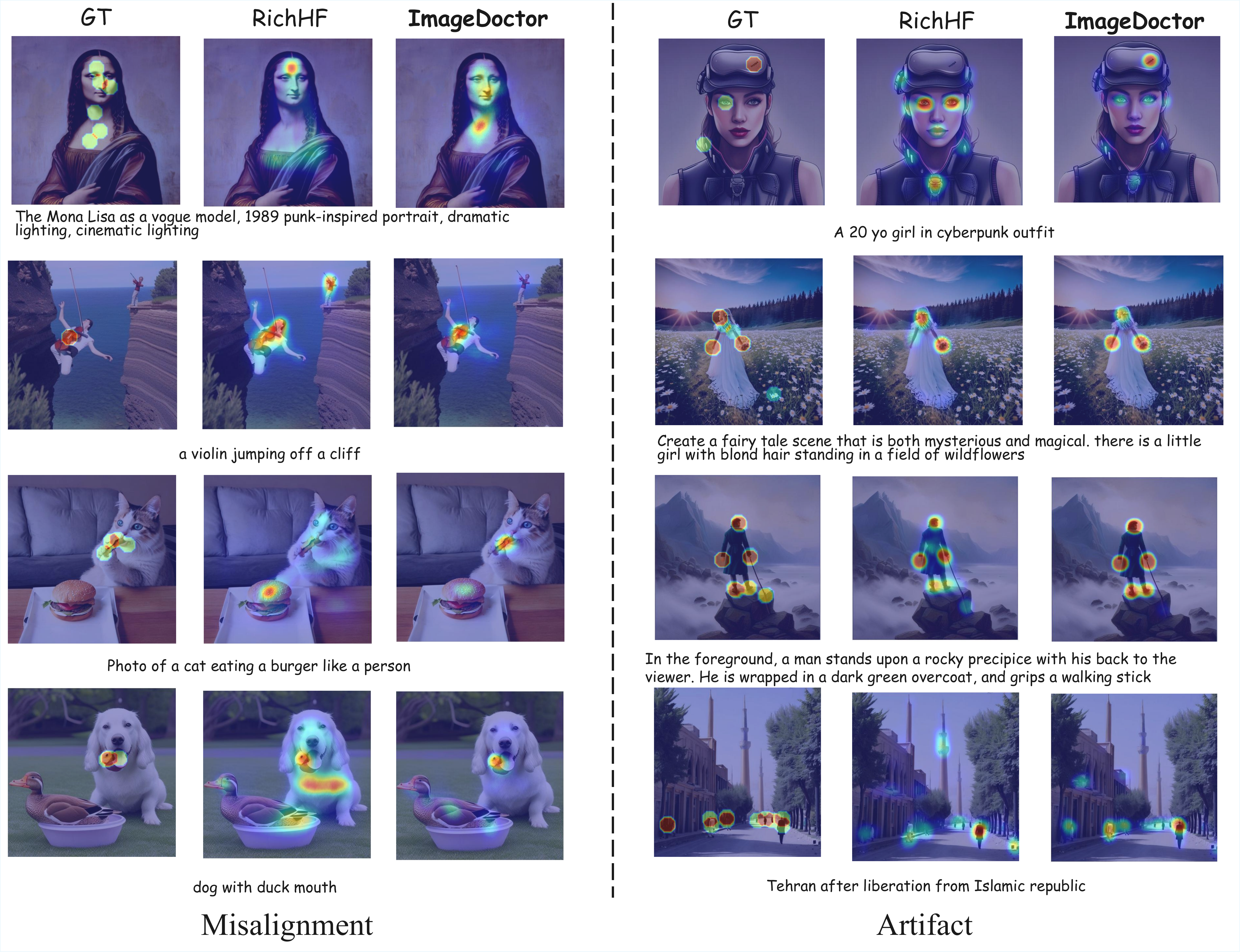}
    \caption{Additional heatmap visualization.}
    \label{fig:add_heatmap}
\end{figure}

As shown in~\cref{fig:add_heatmap}, we provide additional heatmaps for qualitative comparison.

\subsection{Additional Dense-GRPO Results}
In~\cref{fig:add_heatmap}, we provide additional comparison between Flow-GRPO refined with PickScore and DenseFlow-GRPO refined by ImageDoctor. 
\footnotetext[1]{Test using provided checkpoint.}
\begin{figure}
    \centering
    \includegraphics[width=\linewidth]{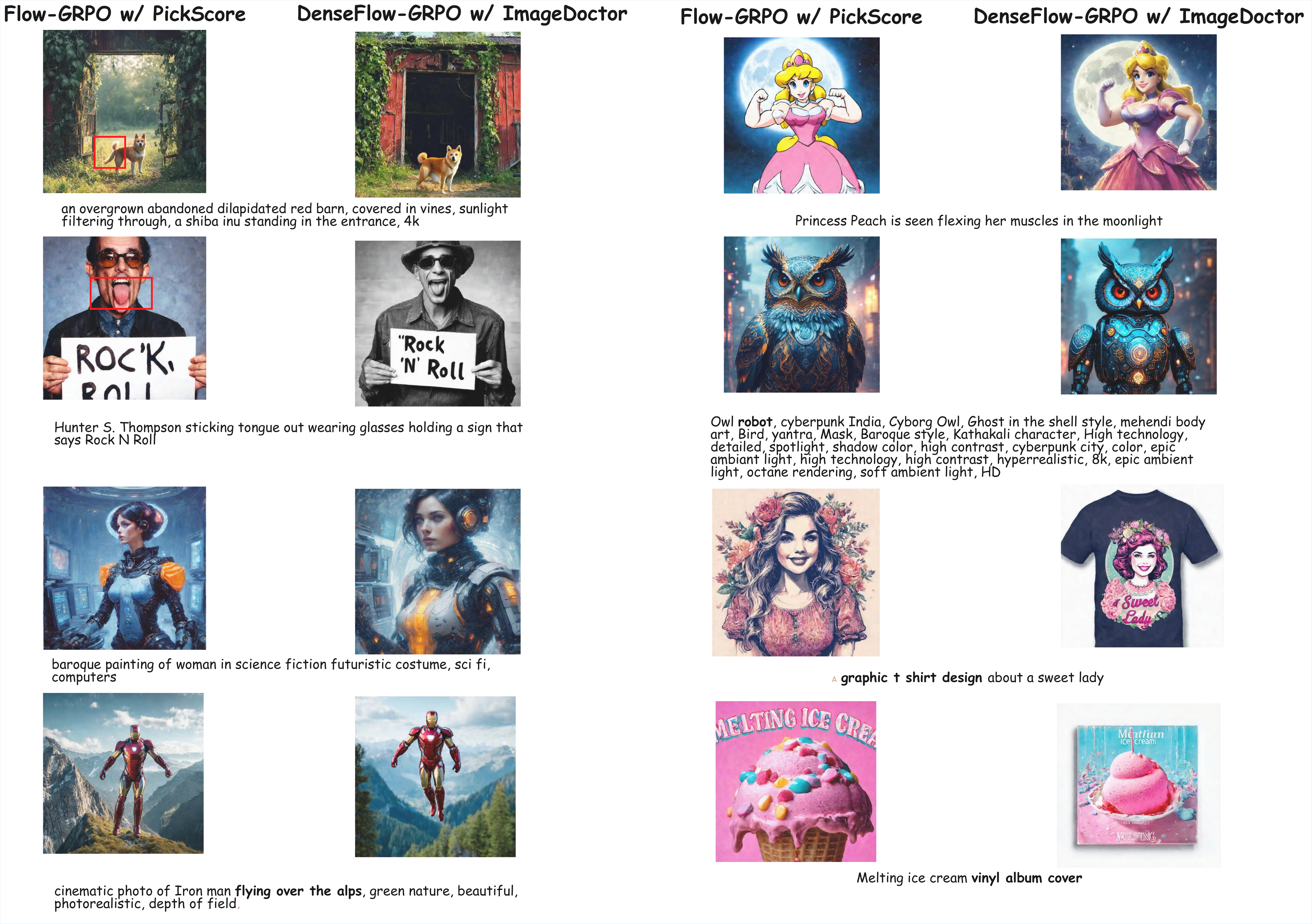}
    \caption{Additional DenseFlow-GRPO visualization.}
    \label{fig:add_dense}
\end{figure}

\section{Limitations}

While ImageDoctor demonstrates strong capability in providing interpretable multi-aspect scoring with spatially grounded feedback, it is important to acknowledge certain limitations that may affect its generalizability and applicability.

\textbf{Challenge of large-scale fine-grained annotation.} Collecting detailed annotations—including multi-aspect scores and heatmaps—is time-consuming and labor-intensive, which limits the volume of available data. The dataset we trained on, i.e. RichHF-18K, is a subset of Pick-a-Pic, which is mostly generated by some old image generation models, \eg  Stable Diffusion XL. This constraint in both scale and recency can restrict the full potential of ImageDoctor. Nevertheless, in this work we show that even with limited and somewhat outdated annotations, ImageDoctor can be effectively trained and still provide valuable dense feedback as a verifier, a reward function, and a metric.

\textbf{Human preference is subjective.} Quantifying image quality is inherently challenging because different people may perceive the same image in very different ways, making it difficult to establish a universally agreed-upon standard. This subjectivity often results in inconsistent annotations, which can affect dataset quality. It also affects heatmap annotations: for instance, annotators may disagree on the exact regions that constitute misalignment, leading to noisy supervision. Consequently, ImageDoctor achieves lower performance on misalignment heatmaps compared to artifact heatmaps.

\section{Prompt for ImageDoctor}
The prompt to ask ImageDoctor to multi-aspect scores and spatial feedback with reasoning is as following:

\begin{tcolorbox}[breakable,colback=orange!5!white, colframe=orange!80!black, title=Prompt for ImageDoctor for T2I Evaluation]
\texttt{<image>\\
Given a caption and an image generated based on this caption, please analyze the provided image in detail. Evaluate it on various dimensions including Semantic Alignment (How well the image content corresponds to the caption), Aesthetics (composition, color usage, and overall artistic quality), Plausibility (realism and attention to detail), and Overall Impression (General subjective assessment of the image's quality). For each evaluation dimension, provide a score between 0-1 and provide a concise rationale for the score. Use a chain-of-thought process to detail your reasoning steps, and enclose all potential important areas and detailed reasoning within <think> and </think> tags. The important areas are represented in following format:” I need to focus on the bounding box area. Proposed regions (xyxy): ..., which is an enumerated list in the exact format:1.[x1,y1,x2,y2];2.[x1,y1,x2,y2];3.[x1,y1,x2,y2]… Here, x1,y1 is the top-left corner, and x2,y2 is the bottom-right corner. Then, within the <answer> and </answer> tags, summarize your assessment in the following format:\\
"Semantic Alignment score: ... \\
Aesthetic score: ...\\
Plausibility score: ...\\
Overall Impression score: ...\\
Misalignment Locations: ...\\
Artifact Locations: ..."\\
No additional text is allowed in the answer section.\\
Your actual evaluation should be based on the quality of the provided image.\\
Your task is provided as follows:\\
Text Caption: [PROMPT]
}
\end{tcolorbox}

\section{LLM Usage Statement}

We employed Gemini-2.5 Flash for preparing reasoning path generation and ChatGPT5 to refine sentence structure and enhance the readability of the manuscript. In addition, Nano Banana was used to assist in generating illustrative figures for clearer presentation. The LLMs were not involved in research ideation or experimental design. LLM assistance on language editing did not influence the substance of the work.

\end{document}